\documentclass[11pt]{article}

\usepackage[T1]{fontenc}
\usepackage[utf8]{inputenc}
\usepackage{lmodern}
\usepackage[margin=1in]{geometry}
\usepackage{microtype}
\usepackage{authblk}

\usepackage{amsmath,amssymb,amsthm,mathtools,mathrsfs,bm}
\usepackage{bbm}
\usepackage{enumitem}
\usepackage{algorithm}
\usepackage{algorithmic}
\usepackage{array}
\usepackage{booktabs}
\usepackage{graphicx}
\usepackage{placeins}
\usepackage[title]{appendix}
\usepackage{natbib}
\usepackage[hidelinks]{hyperref}
\usepackage[nameinlink,capitalize,noabbrev]{cleveref}

\hypersetup{
  pdftitle={Self-Normalized Inference for Constant-Stepsize
    Temporal-Difference Learning under Markovian Sampling},
  pdfauthor={Min Zeng, Yichen Zhang, and Xiaofeng Shao},
  pdfkeywords={temporal-difference learning, self-normalization,
    Markovian sampling, constant stepsize, functional central limit theorem}
}
\theoremstyle{plain}
\newtheorem{theorem}{Theorem}
\newtheorem{lemma}[theorem]{Lemma}
\newtheorem{proposition}[theorem]{Proposition}
\newtheorem{corollary}[theorem]{Corollary}

\theoremstyle{remark}

\theoremstyle{plain}
\newtheorem{assumption}{Assumption}
\crefname{theorem}{theorem}{theorems}
\Crefname{theorem}{Theorem}{Theorems}
\crefname{assumption}{assumption}{assumptions}
\Crefname{assumption}{Assumption}{Assumptions}
\crefname{proposition}{proposition}{propositions}
\Crefname{proposition}{Proposition}{Propositions}
\crefname{corollary}{corollary}{corollaries}
\Crefname{corollary}{Corollary}{Corollaries}
\crefname{lemma}{lemma}{lemmas}
\Crefname{lemma}{Lemma}{Lemmas}
\crefname{remark}{remark}{remarks}
\Crefname{remark}{Remark}{Remarks}

\newenvironment{keywords}
  {\begin{quote}\small\noindent\textbf{Keywords:}\ }
  {\end{quote}}
\newcommand{\acks}[1]{\section*{Acknowledgments and Disclosure of Funding}#1}

\newcommand{\E}{\mathbb E}

\newcommand{\R}{\mathbb R}
\newcommand{\calF}{\mathcal F}
\newcommand{\calG}{\mathcal G}
\newcommand{\calA}{\mathcal A}
\newcommand{\calB}{\mathcal B}
\newcommand{\barA}{\bar A}
\newcommand{\barb}{\bar b}
\newcommand{\tA}{\widetilde A}

\newcommand{\TV}{\mathrm{TV}}
\newcommand{\dto}{\Rightarrow}
\newcommand{\pto}{\xrightarrow{p}}
\newcommand{\norm}[1]{\left\lVert #1\right\rVert}
\newcommand{\Id}{I}

\title{Self-Normalized Inference for Constant-Stepsize
Temporal-Difference Learning under Markovian Sampling}

\author[1]{Min Zeng}
\author[2]{Yichen Zhang}
\author[3,4]{Xiaofeng Shao}
\affil[1]{Department of Biostatistics, City University of Hong Kong,
  Hong Kong SAR, China\\\texttt{zeng.min@my.cityu.edu.hk}}
\affil[2]{Department of Quantitative Methods, Purdue University,
  West Lafayette, Indiana, USA\\\texttt{yichen@purdue.edu}}
\affil[3]{Department of Statistics and Data Science,
  Washington University in St. Louis, St. Louis, Missouri, USA}
\affil[4]{Department of Economics, Washington University in St. Louis,
  St. Louis, Missouri, USA\\\texttt{shaox@wustl.edu}}
\date{August 2026}

\begin{document}

\maketitle

\begin{abstract}Constant-stepsize temporal-difference (TD) learning is attractive for policy
evaluation, but inference from a single Markov trajectory
must account for serial dependence and a stepsize-dependent stationary target.
For fixed-stepsize linear TD, we establish a functional central limit theorem
whose covariance retains the multiplicative component induced by the random
 TD matrix and the stationary iterate error.  We then derive a joint functional
limit for parallel Richardson--Romberg (RR) recursions driven by the same
trajectory.  A Brownian-bridge self-normalizer yields asymptotically pivotal
confidence regions for prespecified state-value contrasts without estimating
the long-run covariance or selecting a bandwidth or batch length.
For such a contrast, the procedure admits a one-pass implementation whose
memory does not grow with the trajectory length.  At a fixed stepsize, the
inferential center is the RR stationary target.  We also study
horizon-indexed designs in which the stepsize remains constant within each run
and decreases across longer horizons.  Under an explicit RR-dependent rate window, 
the residual RR target shift, multiplicative remainder, and initialization effect
are negligible at the root-$n$ scale, yielding inference for the projected
Bellman solution.  Experiments on FrozenLake and Garnet illustrate
stationary-target coverage, RR target correction, and the finite-sample
behavior of the horizon-indexed design.
\end{abstract}
 
\begin{keywords}
temporal-difference learning, self-normalization, Markovian sampling,
constant stepsize, functional central limit theorem
\end{keywords}

\section{Introduction}
\label{sec:intro}

\subsection{Overview and Inferential Targets}

Policy evaluation asks how much future reward a fixed decision rule will
yield from data collected along a trajectory.  It is a basic component of
policy iteration and is also needed when learned policies are compared,
monitored, or deployed \citep{SuttonBarto2018,DannNeumannPeters2014}.  With a
large state space, temporal-difference (TD) learning provides a
streaming estimator of a projected value function
\citep{Sutton1988,TsitsiklisVanRoy1997}.  Standard convergence and
estimation-error analyses, however, do not by themselves yield confidence
statements for the approximate value of a state.

The first difficulty is serial dependence.  Successive TD updates use
transitions from one Markov trajectory, and at a fixed stepsize their
fluctuations are not driven by additive noise alone.  The exact stochastic
driving process also contains a multiplicative term involving the random TD
matrix and the current iterate error.  At a fixed stepsize, this term is
generally part of the first-order path law and must be retained when
characterizing the long-run covariance of the path limit.

The second difficulty concerns the inferential center.  For a sufficiently
small constant stepsize $\alpha$, the recursion has a stationary law with mean
$\theta_\alpha$, which need not equal the projected Bellman solution
$\theta^*$.  A confidence region can therefore have valid coverage for
$\theta_\alpha$ while failing to cover $\theta^*$.  Accounting for sampling
fluctuation and correcting the target are therefore distinct statistical
tasks.

We separate the two tasks by running Richardson--Romberg (RR) recursions on
the same trajectory.  Let $\bar\vartheta_n$ be the average of their weighted
combination and let $\theta_{\mathrm{RR},\alpha}$ be its stationary mean.  The
organizing decomposition is
\begin{equation}
    \bar\vartheta_n-\theta^*
    =
    \bigl(\bar\vartheta_n-\theta_{\mathrm{RR},\alpha}\bigr)
    +
    \bigl(\theta_{\mathrm{RR},\alpha}-\theta^*\bigr).
    \label{eq:intro-target-decomposition}
\end{equation}
The first term is sampling fluctuation around the RR stationary target,
whereas the second is the residual RR target shift. We address the first
term with a Brownian-bridge self-normalizer constructed from the observed
weighted RR path. The resulting statistic is asymptotically pivotal,
requires neither separate long-run covariance estimation nor bandwidth or
batch-length selection, and, for a prespecified contrast, admits a one-pass
online implementation whose memory does not grow with the trajectory length.
RR extrapolation and the horizon-indexed design described below address the
second term.

\Cref{tab:two-inferential-regimes} summarizes the distinction.
\begin{table}[htbp]
\centering
\small
\setlength{\tabcolsep}{4pt}
\begin{tabular}{@{}llll@{}}
\toprule
Design & Stepsize within a run & FCLT center & Leading stochastic term \\
\midrule
Fixed stepsize
& $\alpha_\ell=a_\ell\alpha$
& $\theta_{\mathrm{RR},\alpha}$
& additive and multiplicative TD noise \\
Horizon-indexed
& $\alpha_{\ell,n}=a_\ell\alpha_n$
& $\theta^*$
& additive Markov noise \\
\bottomrule
\end{tabular}
\caption{Two inferential regimes.}
\label{tab:two-inferential-regimes}
\end{table}
The two rows support different practical workflows.  For a learner that
continues to operate at a fixed learning rate, inference concerns its RR
stationary target.  If the projected Bellman solution is instead the target, the terminal
horizon, constant-within-run stepsize, RR design, and TD-parameter
initialization treatment must be specified before the run.

The fixed-stepsize theory requires stability of the mean update matrix but
does not require every random TD update to contract. The joint path law
retains both the multiplicative TD term and the dependence induced by using
the same trajectory at all RR levels. This joint FCLT provides the
theoretical basis for the self-normalized confidence region described above.

The horizon-indexed analysis is a separate triangular-array argument, not a
result obtained by substituting $\alpha_n$ into the fixed-stepsize FCLT.  Across separately planned
runs, $\alpha_n$ decreases with the terminal horizon but remains constant
within each run.  The rate conditions make the residual RR target shift, the
multiplicative remainder, and the initialization effect lower order, changing the center to
$\theta^*$ and the leading process to additive Markov noise.

\subsection{Contributions}

The paper makes three contributions.
\begin{enumerate}[label=(C\arabic*),leftmargin=*]
\item We develop fixed-stepsize path theory for linear TD
under Markov sampling.
The block-forgetting and pullback results construct the stationary recursion
(\Cref{thm:block-gmc,thm:stationary-solution}), while the augmented-chain
decomposition yields its FCLT (\Cref{prop:augmented-poisson};
\Cref{thm:fixed-alpha-fclt}).
\item We develop self-normalized inference for same-trajectory RR
recursions. A joint FCLT preserves their cross-level dependence, and a
Brownian-bridge self-normalizer yields confidence regions for the RR
stationary target without long-run covariance estimation or
bandwidth/batch-length selection. For a prespecified contrast, the
computation is online and one-pass, with memory independent of the
trajectory length
(\Cref{thm:rr-fclt}; \Cref{cor:random-scaling}).

\item We characterize a horizon-indexed regime in which the leading stochastic
input
becomes additive and the residual RR target shift is negligible at the root-$n$
scale.  This yields self-normalized inference for the projected Bellman
solution over an explicit RR-dependent exponent window (\Cref{thm:theta-star-vanishing-fclt} and \Cref{cor:theta-star-rs}).
\end{enumerate}

 Experiments on FrozenLake and
random Garnet MDPs examine stationary-target coverage, RR target correction,
sensitivity to mixing, batch-rule sensitivity, horizon-indexed inference, and online
computation.

\subsection{Relation to Existing Work}
\label{sec:related-work}

TD learning was introduced by \citet{Sutton1988}, and the theory of linear
function approximation under Markov sampling was established by
\citet{TsitsiklisVanRoy1997}; \citet{DannNeumannPeters2014} survey the broader
policy-evaluation literature.  Finite-time analyses include
\citet{BhandariRussoSingal2021,SrikantYing2019}, while general LSA work
quantifies covariance, dimension, and mixing effects
\citep{MouEtAl2020,MouPananjadyWainwrightBartlett2022}.
\citet{DurmusMoulinesNaumovSamsonovWai2021} provide related Markovian
random-matrix stability results and the moment bound used in our
horizon-indexed analysis.

The closest results differ in their sampling regime, within-run stepsize
schedule, asymptotic center, and inference construction.
\Cref{tab:literature-comparison} records these distinctions.
Under independent sampling, \citet{XieZhang2022} obtain an FCLT
and random-scaling inference.  Their random-scaling construction is based on
the partial-sum path and is a form of self-normalization, the terminology used
here.  Under Markovian sampling,
\citet{HuoChenXie2024AAAI,HuoChenXie2022Bias} study constant-stepsize LSA,
including stationary bias, RR reduction, an averaged-iterate CLT, and
long-run-covariance-based inference.  Constant-stepsize Markovian bias and RR error
analyses are further developed by \citet{AllmeierGast2024} and
\citet{LevinNaumovSamsonov2025}.  For i.i.d.\ SGD,
\citet{LiLouRichterWu2024} develop constant-stepsize path theory, online
long-run covariance estimation, and RR extrapolation, whereas
\citet{LiLouSchmidtHieberWu2025} study stationarity, moments, and
high-dimensional concentration.  For nonlinear SA with stepsizes that
decrease within a run, \citet{LiLiangZhang2023} construct a broader class of
self-normalized pivotal statistics from continuous scale-invariant
functionals.

\begin{table}[t]
\centering
\footnotesize
\setlength{\tabcolsep}{1.7pt}
\renewcommand{\arraystretch}{1.02}
\begin{tabular}{@{}>{\raggedright\arraybackslash}p{0.16\linewidth}
                    >{\raggedright\arraybackslash}p{0.10\linewidth}
                    >{\raggedright\arraybackslash}p{0.15\linewidth}
                    >{\raggedright\arraybackslash}p{0.18\linewidth}
                    >{\raggedright\arraybackslash}p{0.11\linewidth}
                    >{\raggedright\arraybackslash}p{0.20\linewidth}@{}}
\toprule
Work & Sampling & Stepsize & Asymptotic center & CLT/FCLT & Inference construction \\
\midrule
\citet{XieZhang2022}
& i.i.d.
& Fixed
& $\theta_\alpha$
& FCLT
& Self-normalized \\

\citet{HuoChenXie2024AAAI,HuoChenXie2022Bias}
& Markovian
& Fixed
& $\theta_{\mathrm{RR},\alpha}$
& CLT
& LRC estimation \\

\citet{LiLouRichterWu2024}
& i.i.d.
& Fixed
& $\theta_{\mathrm{RR},\alpha}$
& FCLT
& LRC estimation \\

\citet{HadaviMouSamsonovWai2026}
& Markovian
& Fixed
& $\theta_\alpha$
& CLT
& -- \\

\citet{LiLiangZhang2023}
& Markovian
& Decreasing within run
& $\theta^\star$
& FCLT
& Self-normalized \\

This paper
& Markovian
& Fixed / horizon-indexed
& $\theta_{\mathrm{RR},\alpha}$ / $\theta^\star$
& FCLT
& Self-normalized \\
\bottomrule
\end{tabular}
\caption{Closest asymptotic inference results.  The asymptotic-center column uses
common notation: $\theta^\star$ is the projected Bellman solution,
$\theta_\alpha$ the fixed-stepsize stationary mean, and
$\theta_{\mathrm{RR},\alpha}$ the RR stationary target.  RR and LRC denote
Richardson--Romberg extrapolation and long-run covariance.}
\label{tab:literature-comparison}
\end{table}

Other Markovian inference regimes are complementary.
\citet{SamsonovSheshukovaMoulinesNaumov2025} give Gaussian approximation and
multiplier block-bootstrap inference for Polyak--Ruppert averaged LSA under
decreasing stepsizes, and \citet{HadaviMouSamsonovWai2026} allow nonlinear
updates and iterate-dependent Markov kernels without developing an inference
procedure.  \citet{LauandMeyn2024} analyze within-run polynomially decreasing
stepsizes under Markovian noise, including a distinct bias regime when the
exponent is at most one half.  \citet{WuLiWeiRinaldo2026}
study Polyak--Ruppert averaged TD under
independent sampling, with polynomially decreasing within-run stepsizes and
plug-in covariance inference.  \citet{WuWeiRinaldo2026} treat Markov sampling
and obtain Gaussian approximations for polynomially decreasing stepsizes.
Our setting instead keeps the stepsize constant within a run and uses a
Brownian bridge to remove the unknown covariance from the limiting quadratic form
\citep{KieferVogelsangBunzel2000,Shao2010,LeeEtAl2022}.

The theory is fixed-dimensional and concerns linear policy evaluation under a
stationary Markov law.  \Cref{sec:formulation,sec:assumptions}
introduce the targets and assumptions; \Cref{sec:main-results,sec:algorithm}
give the fixed-stepsize theory and online implementation of self-normalized
inference;
\Cref{sec:theta-star} studies projected Bellman inference;
\Cref{sec:experiments} presents the numerical experiments; and
\Cref{sec:conclusion} discusses their implications and the scope of the
theory.  Complete proofs are in Online Appendix 1, while experimental
protocols and additional results are in Online Appendix 2.
 \section{Policy Evaluation and Inferential Targets}
\label{sec:formulation}

Three targets play different roles in the analysis.  The parameter
$\theta^*$ is the projected Bellman solution in the linear TD specialization.
A recursion run with a fixed stepsize has stationary mean
$\theta_\alpha$, and an RR combination driven by the same trajectory has stationary target
$\theta_{\mathrm{RR},\alpha}$.  Fixed-stepsize limit theory is centered at
the relevant stationary target; direct inference for $\theta^*$ requires
the additional horizon-indexed regime in \Cref{sec:theta-star}.

\subsection{Markov Decision Processes and Policy Evaluation}

Consider a discounted Markov decision process and a fixed stationary policy
$\pi$.  Once $\pi$ is fixed, the state sequence is a Markov chain with
transition kernel $P^\pi$; write $d^\pi$ for its stationary distribution.  Its
value function is
\[
    V^\pi(s)
    :=
    \E_\pi\!\left[
        \sum_{k=0}^{\infty}\gamma^k R_{t+k+1}
        \,\middle|\, S_t=s
    \right],
    \qquad 0\le\gamma<1,
\]
and satisfies the Bellman equation
\[
    V^\pi = T^\pi V^\pi,
    \qquad
    (T^\pi V)(s)
    =
    \E_\pi[R_{t+1}+\gamma V(S_{t+1})\mid S_t=s].
\]
TD learning exploits this one-step identity: it replaces the unobserved value
on the right-hand side by the current approximation and updates immediately
after observing a transition \citep{Sutton1988,SuttonBarto2018}.

Let $\varphi:\mathcal S\to\R^d$ be a fixed feature map and approximate the
value function by $V_\theta(s)=\varphi(s)^\top\theta$.  For a transition
$(S_t,R_{t+1},S_{t+1})$, define the TD error
\[
    \delta_{t+1}(\theta)
    :=
    R_{t+1}
    +\gamma\varphi(S_{t+1})^\top\theta
    -\varphi(S_t)^\top\theta.
\]
The on-policy linear TD(0) update is
\begin{align}
    \theta_{t+1}^{\alpha}
    &=
    \theta_t^\alpha
    +
    \alpha\varphi(S_t)\delta_{t+1}(\theta_t^\alpha) \notag\\
    &=
    (\Id-\alpha A_{t+1})\theta_t^\alpha+\alpha b_{t+1},
    \label{eq:td-recursion}
\end{align}
where
\begin{equation}
    A_{t+1}
    :=
    \varphi(S_t)
    \bigl(\varphi(S_t)-\gamma\varphi(S_{t+1})\bigr)^\top,
    \qquad
    b_{t+1}
    :=
    R_{t+1}\varphi(S_t).
    \label{eq:td-one-step-matrices}
\end{equation}
Under stationarity and the conditions below, the projected Bellman parameter
solves
\[
    \E[A_{t+1}]\theta^*=\E[b_{t+1}].
\]
Let $\Pi_{d^\pi}$ be the $L^2(d^\pi)$ orthogonal projection onto the linear
feature span.  The population equation is equivalently the projected Bellman
fixed-point equation \citep{TsitsiklisVanRoy1997}
\[
    V_{\theta^*}
    =
    \Pi_{d^\pi}T^\pi V_{\theta^*}.
\]
The approximation $V_{\theta^*}$ need not equal $V^\pi$ unless the value
function is representable by the chosen features.  Throughout the TD
specialization, $\theta^*$ denotes the projected Bellman solution;
approximation error relative to
the unrestricted $V^\pi$ is a separate representation issue and is not
included in the confidence statements below.

We work with a representation that covers \eqref{eq:td-recursion} and related
bounded linear policy-evaluation updates.  Let
$\{Y_t\}_{t\in\mathbb Z}$ be a stationary Markov chain with transition kernel
$P$ and invariant distribution $\mu$.  The update state $Y_t$ may contain the
environment state together with the reward, successor state, eligibility
variables, or other information required for one update.  Let
\[
    A_t=\calA(Y_t)\in\R^{d\times d},
    \qquad
    b_t=\calB(Y_t)\in\R^d.
\]
The update is \eqref{eq:td-recursion}.  In the on-policy example above,
$Y_{t+1}$ contains $(S_t,R_{t+1},S_{t+1})$ and \eqref{eq:td-one-step-matrices}
defines $\calA$ and $\calB$.

Define
\[
    \barA:=\E_\mu[A_t],
    \qquad
    \barb:=\E_\mu[b_t].
\]
The target $\theta^*$ is the unique solution of
$\barA\theta=\barb$, namely $\theta^*=\barA^{-1}\barb$.
\Cref{ass:hurwitz} guarantees that it is well defined.
With
\[
    e_t^\alpha:=\theta_t^\alpha-\theta^*,
    \qquad
    \xi_t:=b_t-A_t\theta^*,
    \qquad
    \tA_t:=A_t-\barA,
\]
the centered recursion is
\begin{equation}
    e_{t+1}^{\alpha}
    =
    (\Id-\alpha\barA)e_t^\alpha
    +
    \alpha\xi_{t+1}
    -
    \alpha\tA_{t+1}e_t^\alpha .
    \label{eq:centered-recursion}
\end{equation}

For integers $a,b$, write
\begin{equation}
    \Pi_{a:b}^{\alpha}
    :=
    (\Id-\alpha A_b)(\Id-\alpha A_{b-1})\cdots(\Id-\alpha A_a),
    \qquad
    \Pi_{a:b}^{\alpha}:=\Id\quad\text{if }a>b .
    \label{eq:product-notation}
\end{equation}

\subsection{Stationary Targets and Richardson--Romberg Extrapolation}

For sufficiently small fixed $\alpha$,
\Cref{thm:stationary-solution} constructs the stationary solution
$\{\theta_t^{\alpha,\circ}\}_{t\in\mathbb Z}$ of
\eqref{eq:td-recursion}.  Its stationary mean is
\[
    \theta_\alpha:=\E[\theta_t^{\alpha,\circ}].
\]
This distinction is central: the fixed-$\alpha$ limit theory is centered at
$\theta_\alpha$, whereas the target solution remains $\theta^*$.

To distinguish the level-specific stationary means from their RR combination,
let $a_\ell>0$ denote fixed RR nodes and $w_\ell$ the corresponding
extrapolation weights.  The node $a_\ell$ is the multiplier defining the
level-specific stepsize $\alpha_\ell=a_\ell\alpha$ of the $\ell$th recursion;
we call $\alpha$ the base stepsize.  We assume that these level-specific
stepsizes satisfy \Cref{ass:stepsize} and run the parallel TD recursions
\begin{equation}
    \theta_{t+1}^{(\ell)}
    =
    (\Id-a_\ell\alpha A_{t+1})\theta_t^{(\ell)}
    +
    a_\ell\alpha b_{t+1},
    \qquad
    \ell=1,\ldots,L .
    \label{eq:parallel-td}
\end{equation}
Let $\theta_t^{a_\ell\alpha,\circ}$ be the stationary version of the $\ell$th recursion and let $\theta_{a_\ell\alpha}$ be its stationary mean.  Given weights $w_1,\ldots,w_L$, define
\begin{equation}
    \vartheta_t^\alpha
    :=
    \sum_{\ell=1}^L w_\ell\theta_t^{(\ell)},
    \qquad
    \vartheta_t^{\alpha,\circ}
    :=
    \sum_{\ell=1}^L w_\ell\theta_t^{a_\ell\alpha,\circ},
    \qquad
    \theta_{\mathrm{RR},\alpha}
    :=
    \E[\vartheta_t^{\alpha,\circ}]
    =
    \sum_{\ell=1}^Lw_\ell\theta_{a_\ell\alpha}.
    \label{eq:rr-target}
\end{equation}
Thus $\theta_{a_\ell\alpha}$ is the stationary mean at level $\ell$,
$\vartheta_t^\alpha$ is the same-trajectory RR combination, and
$\theta_{\mathrm{RR},\alpha}$ is its stationary mean.

For a fixed cancellation order $q_{\mathrm{RR}}\ge1$, the RR design satisfies
\begin{equation}
    \sum_{\ell=1}^L w_\ell=1,
    \qquad
    \sum_{\ell=1}^L w_\ell a_\ell^j=0,
    \quad j=1,\ldots,q_{\mathrm{RR}}.
    \label{eq:rr-weights}
\end{equation}
With $L=q_{\mathrm{RR}}+1$ distinct positive RR nodes,
\eqref{eq:rr-weights} uniquely determines the usual RR weights.  The theory
also allows any fixed finite design satisfying \eqref{eq:rr-weights}.
For the first-order design used in the experiments,
$(a_1,a_2)=(1,2)$ and $(w_1,w_2)=(2,-1)$, so
$\theta_{\mathrm{RR},\alpha}=2\theta_\alpha-\theta_{2\alpha}$.

All levels use the same observed transitions.  The cancellation equations
remove low-order terms from the
constant-stepsize stationary-bias expansion and yield a higher-order bias
upper bound
\citep{HuoChenXie2024AAAI,HuoChenXie2022Bias}.
For a fixed base stepsize, the RR-FCLT in \Cref{thm:rr-fclt} is centered at
$\theta_{\mathrm{RR},\alpha}$ rather than at $\theta^*$.  Equation
\eqref{eq:intro-target-decomposition} separates the resulting target
shift from sampling uncertainty.

\subsection{Partial-Sum Path and Self-Normalizer}

The RR partial-sum process is
\begin{equation}
    C_{n,\mathrm{RR}}(r)
    :=
    n^{-1/2}
    \sum_{t=1}^{\lfloor nr\rfloor}
    (\vartheta_t^\alpha-\theta_{\mathrm{RR},\alpha}),
    \qquad r\in[0,1].
    \label{eq:rr-partial-sum}
\end{equation}
Let
\[
    S_s:=\sum_{t=1}^s\vartheta_t^\alpha,
    \qquad
    \bar\vartheta_n:=S_n/n.
\]
The Brownian-bridge self-normalizer is
\begin{equation}
    \widehat V_n^{\mathrm{SN}}
    :=
    \frac1{n^2}
    \sum_{s=1}^n
    (S_s-s\bar\vartheta_n)(S_s-s\bar\vartheta_n)^\top .
    \label{eq:rs-matrix}
\end{equation}
The centered partial sums form a sample analogue of a Brownian bridge.  In the
weak limit, the endpoint and bridge functional contain the same unknown
covariance factor, which cancels from the quadratic form.  This construction
therefore eliminates the need for a separate long-run covariance estimate.

Here and below, a contrast means a prespecified linear functional
$R\theta$ of the value-function parameter, with
$R\in\R^{q\times d}$.  For example,
$R=\varphi(s_0)^\top$ targets the approximate value of a designated state,
whereas
$R=\{\varphi(s_1)-\varphi(s_2)\}^\top$ targets a difference between two
state values.  Fixing $R$ before the run also permits the one-pass
implementation in \Cref{sec:algorithm} to maintain only the contrast-level
self-normalizer.
 \section{Assumptions for Fixed-Stepsize Inference}
\label{sec:assumptions}

\begin{assumption}[Markov sampling and bounded updates]
\label[assumption]{ass:markov}
\label[assumption]{ass:markov-bounded}
Let $\{Y_t\}_{t\in\mathbb Z}$ be a stationary Markov chain on a Polish
state space $\mathsf Y$.  The chain is irreducible and aperiodic, and it
has transition kernel $P$ and invariant distribution $\mu$ on the Borel
sigma-field.  It is uniformly geometrically ergodic: for constants
$C_0<\infty$ and $\rho\in(0,1)$,
\[
    \sup_{y\in\mathsf Y}\norm{P^k(y,\cdot)-\mu}_{\TV}
    \le C_0\rho^k,
    \qquad k\ge0.
\]
The total-variation convention is
$\|\nu-\mu\|_{\TV}:=\sup_{|f|\le1}|\nu f-\mu f|$.
Moreover, the measurable maps
\[
    \calA:\mathsf Y\to\R^{d\times d},
    \qquad
    \calB:\mathsf Y\to\R^d
\]
satisfy
\[
    \sup_y\norm{\calA(y)}\le K_A,
    \qquad
    \sup_y\norm{\calB(y)}\le K_b .
\]
\end{assumption}

For an irreducible, aperiodic finite-state chain, uniform geometric ergodicity
holds automatically.  In linear TD, bounded features and rewards imply
bounded update coefficients.  These conditions are standard sufficient
assumptions in Markovian LSA analyses; see, for example,
\citet{HuoChenXie2024AAAI}.

\begin{assumption}[Hurwitz mean stability]
\label[assumption]{ass:mean-stability}
\label[assumption]{ass:hurwitz}
The mean matrix $\barA$ is positive stable: every eigenvalue of $\barA$ has
strictly positive real part, or equivalently $-\barA$ is Hurwitz.
\end{assumption}

For standard on-policy TD, \Cref{ass:hurwitz} follows from nonsingularity of
the stationary feature covariance
$G_\varphi:=\E[\varphi(S_t)\varphi(S_t)^\top]$.  Indeed,
$\barA+\barA^\top\succeq2(1-\gamma)G_\varphi$, a stronger coercivity
statement proved in Online Appendix 1.  We retain the
positive-stability formulation because the analysis also covers bounded linear
policy-evaluation recursions whose mean matrix need not be symmetric.

The fixed-stepsize results use the problem-dependent per-recursion stability
threshold $\alpha_{\mathrm{stab},p}$ defined in
\Cref{app:threshold-guide}.  Here the
underlying problem instance comprises the Markov transition law and invariant
distribution, the update maps $\calA$ and $\calB$, and the parameter dimension
$d$;
for TD, the maps encode the feature representation, reward mechanism, and
discount factor.  The threshold is a sufficient stability constant for the
analysis; practical learning-rate selection is a separate task.  Online
Appendix 1 gives its construction as part of the block-contraction proof.

\begin{assumption}[Fixed stepsizes]
\label[assumption]{ass:stepsize}
Fix $p>2$.  The per-recursion stability threshold
$\alpha_{\mathrm{stab},p}$ and the RR base-step threshold
$\alpha_{\mathrm{fix},p}$ are defined in
\Cref{eq:main-alpha-stab,eq:main-alpha-fix}.  Every stepsize $\beta$ used for
an individual recursion in the fixed-stepsize results satisfies
$0<\beta\le\alpha_{\mathrm{stab},p}$.  For an $L$-level RR design with
$\alpha_\ell=a_\ell\alpha$, assume $a_\ell>0$ and
$0<\alpha\le\alpha_{\mathrm{fix},p}$.  This ensures that every RR level uses
an admissible stepsize, $0<a_\ell\alpha\le\alpha_{\mathrm{stab},p}$.
\end{assumption}
 \section{Fixed-Stepsize Limit Theory}
\label{sec:main-results}

We first construct the stationary version of a fixed-stepsize TD recursion and
derive its partial-sum limit around the stationary mean.  The fixed-stepsize
innovation process contains both additive Markov noise and the multiplicative
term involving the TD matrix and iterate error.  Stacking all RR levels gives
the joint limit for the extrapolated estimator, and self-normalization then
removes the unknown long-run covariance from the limiting quadratic form.
\Cref{sec:theta-star} gives the corresponding result for the projected
Bellman solution under a horizon-indexed stepsize design.

\subsection{Stability and the Stationary Recursion}

Although the mean dynamics are stable, the one-step TD update
$\Id-\alpha A_t$ need not be contractive for every transition.  We instead
establish a blockwise forgetting bound that remains valid under Markov
dependence and uniformly over the initial state of the data chain.

\begin{theorem}[Lyapunov block contraction]
\label{thm:block-gmc}
Suppose \Cref{ass:markov-bounded,ass:hurwitz} hold.  For every fixed
$p>2$, the problem-dependent threshold $\alpha_{\mathrm{stab},p}>0$ specified in
\Cref{app:threshold-guide} and constants $c_p,C_p\in(0,\infty)$ can be chosen so that,
for every
$0<\alpha\le\alpha_{\mathrm{stab},p}$, $t\ge0$, starting state $Y_0=y$,
and deterministic $u\in\R^d$,
\begin{equation}
    \E_y\norm{\Pi_{1:t}^{\alpha}u}^p
    \le
    C_p e^{-c_p\alpha t}\norm{u}^p.
    \label{eq:block-gmc}
\end{equation}
The constants are uniform over
$y$, $t$, and $0<\alpha\le\alpha_{\mathrm{stab},p}$, but not over the
dimension or the underlying problem instance.  For two TD recursions
$\{\theta_t^\alpha\}_{t\ge0}$ and
$\{\widetilde\theta_t^\alpha\}_{t\ge0}$ driven by the same Markov path and
started from deterministic vectors $\theta_0,\widetilde\theta_0\in\R^d$,
\[
    \norm{
    \theta_t^\alpha
    -
    \widetilde\theta_t^\alpha
    }_{L^p}
    \le
    C_p e^{-c_p\alpha t}\norm{\theta_0-\widetilde\theta_0}.
\]
\end{theorem}

The bound shows that the recursion forgets its initial iterate on the time
scale $\alpha^{-1}$, without requiring one-step contraction.  This is the
stability input used to construct a stationary recursion and to show that a
deterministic TD initialization contributes only a boundary term to the later
path limits.  Online Appendix 1 proves the bound directly with a Lyapunov
block argument: mean contraction over a sufficiently long mixing block is
combined with uniform control of the within-block fluctuations.  
\begin{theorem}[Pullback stationary TD solution]
\label{thm:stationary-solution}
Suppose \Cref{ass:markov-bounded,ass:hurwitz,ass:stepsize} hold.  For fixed
$\alpha\in(0,\alpha_{\mathrm{stab},p}]$, the pullback limit
\[
    \theta_0^{\alpha,\circ}
    :=
    \lim_{m\to\infty}
    T_0^\alpha\circ T_{-1}^\alpha\circ\cdots\circ T_{-m+1}^\alpha(x),
    \qquad
    T_t^\alpha(\theta):=(\Id-\alpha A_t)\theta+\alpha b_t,
\]
exists in $L^p$, is independent of deterministic $x$, and generates a
stationary solution
\[
    \theta_t^{\alpha,\circ}
    =
    T_t^\alpha(\theta_{t-1}^{\alpha,\circ}).
\]
Moreover, with $\calF_t^Y:=\sigma(Y_s:s\le t)$,
$\theta_t^{\alpha,\circ}$ is $\calF_t^Y$-measurable and has finite $p$th
moment.  In fact,
\[
    \sup_{0<\alpha\le\alpha_{\mathrm{stab},p}}
    \norm{\theta_t^{\alpha,\circ}}_{L^p}<\infty,
\]
the solution is unique among solutions with finite $p$th moment that are
jointly stationary with the two-sided data process, driven by the same data
path, and adapted to
$\{\calF_t^Y\}_{t\in\mathbb Z}$.  In addition, every TD recursion
$\{\theta_t^\alpha\}_{t\ge0}$ started from a deterministic
$\theta_0\in\R^d$ satisfies
\begin{equation}
    \norm{\theta_t^\alpha-\theta_t^{\alpha,\circ}}_{L^p}
    \le
    C e^{-c\alpha t}(1+\norm{\theta_0}),
    \label{eq:stationary-forgetting}
\end{equation}
where $c$ and $C$ can be chosen uniformly over
$0<\alpha\le\alpha_{\mathrm{stab},p}$.
\end{theorem}

The pullback solution provides the exact stationary law around which a
constant-stepsize trajectory fluctuates.  Equation
\eqref{eq:stationary-forgetting} also transfers any path limit proved for the
stationary recursion to a recursion started from a fixed deterministic TD
parameter.  We therefore derive the stationary limit first;
\Cref{thm:fixed-alpha-fclt} states the stationary FCLT and its
deterministic-initialization extension together.

\subsection{Fixed-\texorpdfstring{$\alpha$}{alpha} FCLT}

To distinguish the stationary error process from the arbitrarily initialized
error $e_t^\alpha$ in \eqref{eq:centered-recursion}, define
\[
    h_t^\alpha:=\theta_t^{\alpha,\circ}-\theta^*,
    \qquad
    \bar h_\alpha:=\E[h_t^\alpha]=\theta_\alpha-\theta^*.
\]
The fixed-stepsize innovation process is
\begin{equation}
    \phi_t^\alpha
    :=
    \xi_t-\tA_t h_{t-1}^\alpha-\barA\bar h_\alpha.
    \label{eq:phi-def}
\end{equation}
By stationarity of the TD recursion, $\E[\phi_t^\alpha]=0$.  At a fixed
stepsize, both $\xi_t$ and the multiplicative component
$-\tA_t h_{t-1}^\alpha$ enter the effective first-order innovation.  A Poisson equation on the
data chain alone therefore does not capture the resulting dependence, and we
instead define the augmented chain
\begin{equation}
    Z_t^\alpha:=(Y_t,h_{t-1}^\alpha),
    \qquad
    \calG_t^\alpha:=\sigma(Z_s^\alpha:s\le t),
    \label{eq:augmented-chain}
\end{equation}
with transition kernel $P_\alpha$, stationary law $\pi_\alpha$, and
\[
    \phi_\alpha(y,u):=\xi(y)-\tA(y)u-\barA\bar h_\alpha,
    \qquad
    \phi_t^\alpha=\phi_\alpha(Z_t^\alpha).
\]
A standard uniformly ergodic Markov-chain FCLT does not apply directly because
the iterate coordinate is unbounded and forgets its initialization on the
time scale $\alpha^{-1}$.  Online Appendix 1 uses finite-memory
approximation and projective summability to establish the following
decomposition.

\begin{proposition}[Augmented-chain Poisson equation]
\label[proposition]{prop:augmented-poisson}
For fixed $\alpha\in(0,\alpha_{\mathrm{stab},p}]$, there exists
$V_\alpha\in L^2(\pi_\alpha)$ such that
\begin{equation}
    V_\alpha-P_\alpha V_\alpha=\phi_\alpha
    \qquad \text{in }L^2(\pi_\alpha).
    \label{eq:augmented-poisson}
\end{equation}
Define
\begin{equation}
    D_{t+1}^\alpha
    :=
    V_\alpha(Z_{t+1}^\alpha)
    -
    P_\alpha V_\alpha(Z_t^\alpha).
    \label{eq:D-alpha-def}
\end{equation}
Then $\{D_{t+1}^\alpha,\calG_{t+1}^\alpha\}$ is a stationary martingale-difference sequence and
\begin{equation}
    \phi_t^\alpha
    =
    D_{t+1}^\alpha
    +
    V_\alpha(Z_t^\alpha)
    -
    V_\alpha(Z_{t+1}^\alpha).
    \label{eq:phi-martingale-coboundary}
\end{equation}
\end{proposition}

Define
\begin{equation}
    \Sigma_{\phi,\alpha}
    :=
    \E[D_1^\alpha(D_1^\alpha)^\top],
    \qquad
    \Omega_\alpha
    :=
    \barA^{-1}\Sigma_{\phi,\alpha}\barA^{-\top}.
    \label{eq:fixed-alpha-covariance}
\end{equation}

The exact identity used in this reduction is
\begin{equation}
    \sum_{t=1}^{k}(h_t^\alpha-\bar h_\alpha)
    =
    \barA^{-1}\sum_{t=1}^{k}\phi_t^\alpha
    +
    \left(\Id-\alpha^{-1}\barA^{-1}\right)(h_k^\alpha-h_0^\alpha).
    \label{eq:fclt-exact-partial-sum-main}
\end{equation}
Together with \eqref{eq:phi-martingale-coboundary}, it isolates the martingale
that determines the long-run covariance in
\eqref{eq:fixed-alpha-covariance}.

\begin{theorem}[Fixed-\texorpdfstring{$\alpha$}{alpha} TD FCLT]
\label{thm:fixed-alpha-fclt}
Suppose \Cref{ass:markov-bounded,ass:hurwitz,ass:stepsize} hold.
For the stationary TD solution,
\begin{equation}
    n^{-1/2}
    \sum_{t=1}^{\lfloor nr\rfloor}
    (\theta_t^{\alpha,\circ}-\theta_\alpha)
    \dto
    \Omega_\alpha^{1/2}W_d(r)
    \label{eq:stationary-fclt}
\end{equation}
in $D([0,1],\R^d)$, where $W_d$ is a standard $d$-dimensional Brownian
motion.

Moreover, let $\{\theta_t^\alpha\}_{t\ge0}$ be the TD recursion in
\eqref{eq:td-recursion} started from any fixed deterministic vector
$\theta_0\in\R^d$.  Then
\[
    n^{-1/2}
    \sum_{t=1}^{\lfloor nr\rfloor}
    (\theta_t^\alpha-\theta_\alpha)
    \dto
    \Omega_\alpha^{1/2}W_d(r)
\]
in $D([0,1],\R^d)$.
\end{theorem}

The limit is centered at $\theta_\alpha$, not at $\theta^*$.  Its covariance
is generated by the fixed-stepsize innovation process in
\eqref{eq:phi-def}, so the multiplicative term remains in the root-$n$ path
limit before self-normalization and cannot generally be replaced by additive
Markov noise.
The Brownian-bridge statistic removes this covariance
without estimating it; this is why an FCLT, rather than only an endpoint CLT,
is needed.  The second conclusion extends the same path limit to any fixed
deterministic initialization of the TD parameter.  We next turn to the joint
limit for RR recursions driven by the same trajectory.

\subsection{Same-Trajectory RR Limit and Self-Normalized Inference}

Recall from \Cref{sec:formulation} that a fixed RR design of cancellation order
$q_{\mathrm{RR}}$ satisfies \eqref{eq:rr-weights}.  The results below do not
optimize its RR nodes or weights.

The finite-order stationary-bias expansion of
\citet{HuoChenXie2022Bias} implies that a fixed RR design satisfying
\eqref{eq:rr-weights} obeys
\begin{equation}
    \norm{\theta_{\mathrm{RR},\alpha}-\theta^*}
    \le C_{\mathrm{RR}}\alpha^{q_{\mathrm{RR}}+1}
    \label{eq:rr-bias-exact-order}
\end{equation}
for all sufficiently small admissible $\alpha$.  Online Appendix 1 gives the
full expansion, translates the cited stepsize condition into our notation, and
states the dependence on the fixed cancellation order.  The bound enters when the center is replaced by
$\theta^*$ in \Cref{sec:theta-star}; fixed-$\alpha$ coverage of
$\theta_{\mathrm{RR},\alpha}$ follows without it.  

Define the stacked stationary process
\[
    \Theta_t^{\alpha,\circ}
    :=
    ((\theta_t^{a_1\alpha,\circ})^\top,\ldots,(\theta_t^{a_L\alpha,\circ})^\top)^\top,
    \qquad
    \Theta_\alpha
    :=
    (\theta_{a_1\alpha}^\top,\ldots,\theta_{a_L\alpha}^\top)^\top,
\]
and
\[
    B_{\mathrm{RR}}:=(w_1\Id,\ldots,w_L\Id)\in\R^{d\times Ld}.
\]
Set
$\vartheta_t^{\alpha,\circ}:=B_{\mathrm{RR}}\Theta_t^{\alpha,\circ}$.
Then
$\theta_{\mathrm{RR},\alpha}=B_{\mathrm{RR}}\Theta_\alpha$.

All RR levels are updated using the same transitions, so their partial sums are
generally cross-correlated.  This same-trajectory construction is also used by
\citet{HuoChenXie2024AAAI}.  The RR-FCLT in \Cref{thm:rr-fclt} retains the
resulting cross-level dependence and thereby supports self-normalized inference
for the weighted estimator.

\begin{theorem}[RR-FCLT]
\label{thm:rr-fclt}
Suppose \Cref{ass:markov-bounded,ass:hurwitz,ass:stepsize} hold, and fix a
stepsize $\alpha>0$ such that
$a_\ell\alpha\le\alpha_{\mathrm{stab},p}$ for every
$\ell=1,\ldots,L$.  There exists a positive-semidefinite
long-run covariance matrix $\Omega_{\Theta,\alpha}$ such that, as
$n\to\infty$,
\[
    n^{-1/2}
    \sum_{t=1}^{\lfloor nr\rfloor}
    (\Theta_t^{\alpha,\circ}-\Theta_\alpha)
    \dto
    \Omega_{\Theta,\alpha}^{1/2}W_{Ld}(r)
\]
in $D([0,1],\R^{Ld})$ equipped with the $J_1$ topology.  Consequently,
\begin{equation}
    n^{-1/2}
    \sum_{t=1}^{\lfloor nr\rfloor}
    (\vartheta_t^{\alpha,\circ}-\theta_{\mathrm{RR},\alpha})
    \dto
    \Omega_{\vartheta,\alpha}^{1/2}W_d(r),
    \qquad
    \Omega_{\vartheta,\alpha}
    :=B_{\mathrm{RR}}\Omega_{\Theta,\alpha}B_{\mathrm{RR}}^\top,
    \label{eq:rr-fclt}
\end{equation}
in $D([0,1],\R^d)$ equipped with the $J_1$ topology.

Both convergences above remain valid pointwise when the stationary recursions
are replaced by recursions initialized at any deterministic tuple
$(\theta_0^{(1)},\ldots,\theta_0^{(L)})\in(\R^d)^L$ that is fixed as
$n\to\infty$.
\end{theorem}

The off-diagonal blocks of $\Omega_{\Theta,\alpha}$ represent the dependence
induced by using the same trajectory at all RR levels.  They enter the
weighted variance through
$B_{\mathrm{RR}}\Omega_{\Theta,\alpha}B_{\mathrm{RR}}^\top$, so
independent marginal limits would describe a different experiment.  The
self-normalizer is constructed from the observed weighted RR path and
therefore does not require these cross-level covariance blocks to be estimated
separately.

\begin{corollary}[Self-normalized inference]
\label[corollary]{cor:random-scaling}
\label[corollary]{cor:self-normalized-fixed}
Assume the RR-FCLT in \Cref{thm:rr-fclt} holds at a fixed stepsize
$\alpha>0$.  Fix $q\in\{1,\ldots,d\}$ and a full-row-rank contrast
$R\in\R^{q\times d}$ such that
\[
    \Gamma_{R,\alpha}
    :=
    R\Omega_{\vartheta,\alpha}R^\top
    \succ0.
\]
For $H_0:R\theta_{\mathrm{RR},\alpha}=c$, write
$\widehat V_{n,R}^{\mathrm{SN}}
:=R\widehat V_n^{\mathrm{SN}}R^\top$.  On the event
$\widehat V_{n,R}^{\mathrm{SN}}\succ0$, define
\begin{equation}
    T_n^{\mathrm{SN}}
    :=
    n(R\bar\vartheta_n-c)^\top
    (\widehat V_{n,R}^{\mathrm{SN}})^{-1}
    (R\bar\vartheta_n-c).
    \label{eq:rs-stat}
\end{equation}
Then, under $H_0$, as $n\to\infty$,
\begin{equation}
    T_n^{\mathrm{SN}}
    \dto
    W_q(1)^\top
    \left[
      \int_0^1
      \bar W_q(r)\bar W_q(r)^\top\,dr
    \right]^{-1}
    W_q(1),
    \qquad
    \bar W_q(r)=W_q(r)-rW_q(1).
    \label{eq:rs-limit}
\end{equation}
Here $W_q$ is a standard $q$-dimensional Brownian motion.  The limit is
pivotal and does not involve $\Omega_{\vartheta,\alpha}$ or
$\Gamma_{R,\alpha}$.
\end{corollary}

Let $\kappa_{q,1-\eta}$ denote the $(1-\eta)$ quantile of the Brownian
functional in \Cref{eq:rs-limit}, whose distribution is continuous by
\Cref{lem:main-brownian-bridge} in Appendix B.  On the positive-definite event, the corresponding
confidence region is
\begin{equation}
    \mathcal C_{n,1-\eta}
    :=
    \left\{
       u\in\R^q:
       n(R\bar\vartheta_n-u)^\top
       (\widehat V_{n,R}^{\mathrm{SN}})^{-1}
       (R\bar\vartheta_n-u)
       \le\kappa_{q,1-\eta}
    \right\}.
    \label{eq:self-normalized-confidence-region}
\end{equation}
The probability of this event tends to one, and the region has asymptotic
coverage $1-\eta$ for $R\theta_{\mathrm{RR},\alpha}$; no region is reported on
the complement.
When $q=1$ and $R=r^\top$, it reduces to
\begin{equation}
    r^\top\bar\vartheta_n
    \ \mathbin{\pm}\
    \sqrt{
       \frac{\kappa_{1,1-\eta}}{n}\,
       r^\top\widehat V_n^{\mathrm{SN}}r
    }.
    \label{eq:self-normalized-scalar-interval}
\end{equation}
The critical value depends only on the contrast dimension and nominal level
and can therefore be simulated once from Brownian motion.
\Cref{app:self-normalization-proof} proves the pivotal limit; Online Appendix 2
reports the Brownian simulation used in the experiments.  For the proof,
\Cref{prop:main-sn-transfer} extends the statistic by setting it to zero off
the positive-definite event.  The
procedure therefore requires neither consistent long-run covariance
estimation nor selection of a bandwidth or batch length.
\Cref{sec:algorithm} gives a one-pass recursion for the contrast-level
self-normalizer.
 \section{Online Implementation of Self-Normalized Inference}
\label{sec:algorithm}

For a prespecified contrast $R$, the RR estimator and its Brownian-bridge
self-normalizer can be updated in one pass, with memory independent of the
reporting horizon.  Set
\[
    x_t:=R\vartheta_t\in\R^q,
    \qquad
    U_t:=\sum_{s=1}^t x_s,
    \qquad
    \bar x_t:=U_t/t,
\]
and maintain
\[
    G_t:=\sum_{s=1}^t U_sU_s^\top,
    \qquad
    H_t:=\sum_{s=1}^t sU_s,
    \qquad
    c_t:=\sum_{s=1}^t s^2 .
\]
The self-normalizer for the contrast process $x_t=R\vartheta_t$ is
\begin{equation}
    \widehat V_{t,R}^{\mathrm{SN}}
    :=
    R\widehat V_t^{\mathrm{SN}}R^\top
    =
    t^{-2}\left\{
       G_t-H_t\bar x_t^\top-\bar x_tH_t^\top
       +c_t\bar x_t\bar x_t^\top
    \right\}.
    \label{eq:online-rs}
\end{equation}

Algorithm~\ref{alg:online} uses the reindexed convention in which
$(A_t,b_t)$ updates $\theta_{t-1}$ to $\theta_t$.  This is identical to
\eqref{eq:td-recursion} after shifting the time index by one.

\begin{algorithm}[H]
\caption{Online RR-TD with self-normalized inference}
\label{alg:online}
\begin{algorithmic}[1]
\STATE \textbf{Input:} base stepsize $\alpha$, fixed RR design
$\{(a_\ell,w_\ell)\}_{\ell=1}^L$, contrast $R$, nominal level $1-\eta$,
critical value $\kappa_{q,1-\eta}$, terminal horizon $T$.
\STATE Initialize $\theta_0^{(\ell)}$, $U_0=0$, $G_0=0$, $H_0=0$, and
$c_0=0$.
\FOR{$t=1,\ldots,T$}
    \STATE Observe $Y_t$ and form $A_t,b_t$.
    \FOR{$\ell=1,\ldots,L$}
        \STATE $\theta_t^{(\ell)}
        \leftarrow
        (\Id-a_\ell\alpha A_t)\theta_{t-1}^{(\ell)}
        +a_\ell\alpha b_t$.
    \ENDFOR
    \STATE $\vartheta_t\leftarrow
       \sum_{\ell=1}^Lw_\ell\theta_t^{(\ell)}$;
       $x_t\leftarrow R\vartheta_t$.
    \STATE $U_t\leftarrow U_{t-1}+x_t$;
       $\bar x_t\leftarrow U_t/t$.
    \STATE $G_t\leftarrow G_{t-1}+U_tU_t^\top$;
       $H_t\leftarrow H_{t-1}+tU_t$;
       $c_t\leftarrow c_{t-1}+t^2$.
\ENDFOR
\STATE At $t=T$, compute $\widehat V_{T,R}^{\mathrm{SN}}$ from
\Cref{eq:online-rs} and report the confidence region in
\Cref{eq:self-normalized-confidence-region}.
\end{algorithmic}
\end{algorithm}

Excluding implementation-specific temporary storage for the current update
coefficients, a dense contrast requires $O(Ld+qd+q^2)$ persistent storage.
After the TD matrix--vector products, computing $R\vartheta_t$ costs $O(qd)$
and updating the self-normalizer costs $O(q^2)$ per step.  The $O(qd)$ storage
and computational costs can be reduced when $R$ is sparse or applied
implicitly.  An
implementation should not form an explicit inverse.  All experiments in
Section~\ref{sec:experiments} use scalar contrasts ($q=1$).  For a
matrix-valued contrast, the following describes the recommended
finite-precision implementation rather than an additional empirical result.
The contrast-level self-normalizer is symmetrized, and
the associated linear system is solved using a Cholesky or symmetric
factorization.  If a documented numerical rank
check fails, the interval is reported as unavailable; the accumulation
details, rank tolerance, and failure check are given in Online Appendix 2.

The recursion is online in computation and memory after $R$ is fixed, whereas
the coverage statement is for the prespecified reporting horizon.
 \section{Inference for the Projected Bellman Solution}
\label{sec:theta-star}

\subsection{Horizon-Indexed Design and Remainder Conditions}

The fixed-stepsize theory is exact for the RR stationary target.  To obtain
inference for the projected Bellman solution $\theta^*$, we let the base
stepsize depend on the reporting
horizon while keeping it constant within each run.  In this
regime, three additional terms must be negligible at the root-$n$ scale: the
residual RR target shift, the multiplicative remainder, and the initialization
boundary.  The first is controlled by the RR bias order, whereas the latter
two are controlled by the small-step stationary scale and block forgetting.
Specifically, consider
\[
    \alpha=\alpha_n\downarrow0.
\]
For each terminal horizon $n$, the stepsize is held constant throughout the
trajectory but decreases across the triangular array indexed by $n$.  Let
\begin{equation}
    \alpha_{\ell,n}:=a_\ell\alpha_n,
    \qquad
    \vartheta_{t,n}^{\circ}
    :=
    \sum_{\ell=1}^Lw_\ell
    \theta_t^{\alpha_{\ell,n},\circ},
    \qquad
    \theta_{\mathrm{RR},n}
    :=
    \E[\vartheta_{t,n}^{\circ}]
    =
    \sum_{\ell=1}^Lw_\ell\theta_{\alpha_{\ell,n}}.
    \label{eq:theta-star-triangular-definitions}
\end{equation}

The horizon-indexed result uses the per-recursion moment and bias thresholds
$\alpha_{\mathrm{mom},p}$ and $\alpha_{\mathrm{bias},m}$, together with the
corresponding base-stepsize thresholds $\alpha_{\mathrm{RR},m}$ and
$\alpha_{\mathrm{adm},p,m}$ defined in \Cref{eq:main-alpha-adm}.
Appendix A summarizes these sufficient proof thresholds and their roles.

\begin{assumption}[Horizon-indexed stepsizes]
\label[assumption]{ass:horizon-indexed}
Fix $p>2$ and an RR design of cancellation order $q_{\mathrm{RR}}$ satisfying
\eqref{eq:rr-weights}.  For each terminal horizon $n$, the base stepsize
$\alpha_n$ is held constant throughout the run, and level $\ell$ uses
$\alpha_{\ell,n}=a_\ell\alpha_n$.  Assume that:
\begin{enumerate}[label=(\roman*),leftmargin=2.8em]
\item $\alpha_n\downarrow0$ and
$\alpha_n\le\alpha_{\mathrm{adm},p,q_{\mathrm{RR}}}$ for all sufficiently
large $n$;
\item
\begin{equation}
    n^{1-2/p}\alpha_n\longrightarrow\infty;
    \label{eq:vanishing-boundary-rate}
\end{equation}
\item
\begin{equation}
    \sqrt n\,\alpha_n^{q_{\mathrm{RR}}+1}\longrightarrow0;
    \label{eq:vanishing-bias-rate}
\end{equation}
\item for recursions started from fixed deterministic vectors,
\begin{equation}
    \alpha_n\sqrt n\longrightarrow\infty.
    \label{eq:theta-star-deterministic-rate-main}
\end{equation}
\end{enumerate}
\end{assumption}

Parts (i)--(iii) are used for the stationary triangular-array result.  Part
(iv) is needed only to transfer that result to fixed deterministic TD
initializations; the prespecified TD-burn-in condition in \Cref{sec:rate-window}
is an alternative.

The horizon-indexed result uses a triangular-array argument that tracks the
changing center and stochastic terms across runs.  To expose this change,
define
\[
    h_t^{(\ell,n)}
    :=
    \theta_t^{\alpha_{\ell,n},\circ}-\theta^*,
    \qquad
    \bar h_{\ell,n}:=\E[h_t^{(\ell,n)}].
\]
The fixed-stepsize innovation process for the $\ell$th recursion can be rewritten as
\begin{equation}
    \phi_t^{(\ell,n)}
    =
    \xi_t-\zeta_t^{(\ell,n)},
    \qquad
    \zeta_t^{(\ell,n)}
    :=
    \tA_t h_{t-1}^{(\ell,n)}
    -
    \E[\tA_t h_{t-1}^{(\ell,n)}],
    \label{eq:theta-star-force-decomposition}
\end{equation}
where the stationary mean equation gives
\[
    \E[\tA_t h_{t-1}^{(\ell,n)}]
    =
    -\barA\bar h_{\ell,n}.
\]
Since $\sum_{\ell=1}^Lw_\ell=1$, the RR-combined innovation process is
\begin{equation}
    \sum_{\ell=1}^Lw_\ell\phi_t^{(\ell,n)}
    =
    \xi_t-\zeta_{t,n}^{\mathrm{RR}},
    \qquad
    \zeta_{t,n}^{\mathrm{RR}}
    :=
    \sum_{\ell=1}^Lw_\ell\zeta_t^{(\ell,n)}.
    \label{eq:theta-star-combined-force}
\end{equation}
For fixed $\alpha$, the multiplicative component in
\eqref{eq:theta-star-force-decomposition} enters the leading long-run
covariance.  Under the horizon-indexed conditions in \Cref{ass:horizon-indexed}, its partial-sum
contribution is $o_p(\sqrt n)$, and the limiting driving process reduces to
the additive Markov noise
\[
    \xi_t=b_t-A_t\theta^*.
\]

A uniform small-stepsize moment bound gives
$\|\theta_0^{\alpha,\circ}-\theta^*\|_{L^p}=O(\sqrt{\alpha})$; see Online
Appendix 1.6.1.  This is the scale at which the multiplicative component is shown
to be negligible across the triangular array.  The moment bound controls the
stationary TD boundary and multiplicative
remainder, while \eqref{eq:rr-bias-exact-order} controls the residual RR
target shift.

\begin{table}[t]
\centering
\small
\setlength{\tabcolsep}{4pt}
\begin{tabular}{lll}
\toprule
Term & Scale or control & Sufficient condition \\
\midrule
Residual RR target shift
& $O(\alpha_n^{q_{\mathrm{RR}}+1})$
& $\sqrt n\,\alpha_n^{q_{\mathrm{RR}}+1}\to0$ \\
Stationary TD boundary
& $L^p$ maximal control
& $n^{1-2/p}\alpha_n\to\infty$ \\
Multiplicative component
& Projective maximal control
& \Cref{ass:horizon-indexed}(i)--(ii) \\
Deterministic initialization
& Block-forgetting bound
& $\alpha_n\sqrt n\to\infty$ or prescribed TD burn-in \\
\bottomrule
\end{tabular}
\caption{Sufficient controls for the horizon-indexed argument.}
\label{tab:vanishing-remainders}
\end{table}

\subsection{FCLT and Self-Normalized Inference}

Let $V_\xi$ be the bounded solution of the classical Markov-chain Poisson equation
\[
    V_\xi-PV_\xi=\xi,
\]
and define
\[
    D_{t+1}^{\xi}
    :=
    V_\xi(Y_{t+1})-PV_\xi(Y_t),
    \qquad
    \Sigma_\xi
    :=
    \E[D_1^\xi(D_1^\xi)^\top].
\]
The small-stepsize limiting covariance is
\begin{equation}
    \Omega_0
    :=
    \barA^{-1}\Sigma_\xi\barA^{-\top}.
    \label{eq:theta-star-omega-zero}
\end{equation}

Unlike within-run decreasing-stepsize schemes, each recursion here is
homogeneous within its horizon-indexed run.  The proof must therefore control simultaneously the
residual RR target shift within that run, the $\alpha_n$-dependent TD boundary, and the
multiplicative component.

\begin{theorem}[Horizon-indexed RR-FCLT for
\texorpdfstring{$\theta^*$}{theta*}]
\label{thm:theta-star-vanishing-fclt}
\label{thm:rowwise-fclt}
Suppose \Cref{ass:markov-bounded,ass:hurwitz} and
\Cref{ass:horizon-indexed}(i)--(iii) hold.  Then the stationary
triangular-array process satisfies
\begin{equation}
    n^{-1/2}
    \sum_{t=1}^{\lfloor nr\rfloor}
    \bigl(
        \vartheta_{t,n}^{\circ}-\theta^*
    \bigr)
    \dto
    \Omega_0^{1/2}W_d(r)
    \label{eq:theta-star-main-fclt}
\end{equation}
in $D([0,1],\R^d)$.

Moreover, for each $\ell=1,\ldots,L$, let
$\{\theta_t^{\alpha_{\ell,n}}\}_{t\ge0}$ be the TD recursion in
\eqref{eq:td-recursion} started from a deterministic vector
$\theta_0^{(\ell)}$.  Suppose that
\[
    \max_{1\le\ell\le L}\norm{\theta_0^{(\ell)}}<\infty,
\]
and define
\[
    \vartheta_{t,n}
    :=
    \sum_{\ell=1}^L
    w_\ell
    \theta_t^{\alpha_{\ell,n}}.
\]
If, in addition, \Cref{ass:horizon-indexed}(iv) holds,
then \eqref{eq:theta-star-main-fclt} also holds with
$\vartheta_{t,n}^{\circ}$ replaced by $\vartheta_{t,n}$.
\end{theorem}

The central reduction proved in Online Appendix 1 is
\begin{equation}
    \sup_{0\le r\le1}
    \left\|
        n^{-1/2}
        \sum_{t=1}^{\lfloor nr\rfloor}
        (\vartheta_{t,n}^{\circ}-\theta^*)
        -
        \barA^{-1}
        n^{-1/2}
        \sum_{t=1}^{\lfloor nr\rfloor}
        \xi_t
    \right\|
    \pto0.
    \label{eq:theta-star-main-reduction}
\end{equation}
The three discarded terms are the centered multiplicative component
$\zeta_{t,n}^{\mathrm{RR}}$, the TD boundary term, and the residual RR target shift
$\theta_{\mathrm{RR},n}-\theta^*$.

Online Appendix 1 proves \eqref{eq:theta-star-main-reduction} with constants
that may depend on the fixed underlying problem instance, moment index, and RR
design, but not on
$n$.  The horizon-indexed rates make the multiplicative remainder negligible and
the residual RR target shift smaller than $n^{-1/2}$; consequently, the
additive Markov noise determines $\Omega_0$ and the process is centered at
$\theta^*$.

For the deterministically initialized trajectory in
Theorem~\ref{thm:theta-star-vanishing-fclt}, define
\[
    S_{s,n}:=\sum_{t=1}^s\vartheta_{t,n},
    \qquad
    \bar\vartheta_n:=S_{n,n}/n,
\]
and
\begin{equation}
    \widehat V_{n,*}^{\mathrm{SN}}
    :=
    \frac1{n^2}
    \sum_{s=1}^n
    (S_{s,n}-s\bar\vartheta_n)
    (S_{s,n}-s\bar\vartheta_n)^\top.
    \label{eq:theta-star-rs-matrix-main}
\end{equation}

\begin{corollary}[Self-normalized inference for
\texorpdfstring{$\theta^*$}{theta*}]
\label[corollary]{cor:theta-star-rs}
\label[corollary]{cor:self-normalized-theta-star}
Assume the preceding FCLT for deterministic starts. Fix
$q\in\{1,\ldots,d\}$. Let $R\in\R^{q\times d}$ have full row rank and satisfy
\[
    \Gamma_{R,0}:=R\Omega_0R^\top\succ0.
\]
Applying the self-normalizer in \Cref{cor:random-scaling} to the process at
horizon $n$ gives the same Brownian functional as in \eqref{eq:rs-limit}.
Replacing $\widehat V_n^{\mathrm{SN}}$ by
$\widehat V_{n,*}^{\mathrm{SN}}$ in
\Cref{eq:self-normalized-confidence-region,eq:self-normalized-scalar-interval}
therefore gives confidence regions and scalar intervals with asymptotic
coverage $1-\eta$ for $R\theta^*$.
\end{corollary}

The same Brownian critical values apply, with center
$R\theta_{\mathrm{RR},\alpha}$ in the fixed-stepsize regime and $R\theta^*$
in the horizon-indexed regime.

\subsection{Rate Window and TD Burn-In}
\label{sec:rate-window}

When $\alpha_n=n^{-\nu}$, the RR bias condition requires
$\nu>1/\{2(q_{\mathrm{RR}}+1)\}$, while the stationary-recursion boundary
condition requires $\nu<1-2/p$.  Thus the stationary theorem permits
\[
    \frac1{2(q_{\mathrm{RR}}+1)}
    <
    \nu
    <
    1-\frac2p.
\]
For deterministic initialization without TD burn-in,
\eqref{eq:theta-star-deterministic-rate-main} also requires $\nu<1/2$,
so the range becomes
\[
    \frac1{2(q_{\mathrm{RR}}+1)}
    <
    \nu
    <
    \min\left\{
        \frac12,\,
        1-\frac2p
    \right\}.
\]
For first-order RR, $q_{\mathrm{RR}}=1$ and the lower endpoint is $1/4$.
Higher cancellation order lowers the left endpoint and can create a nonempty
exponent window for direct inference on the projected Bellman solution.
Additional cancellation generally requires more RR nodes and may produce larger
signed weights, increasing finite-sample variability; optimizing this trade-off
is beyond the present scope.  For a proposed exponent $\nu$, one first fixes a moment
index $p>2$ for which the displayed inequalities hold.  Bounded updates allow
such fixed higher moments, although the associated admissible stepsize
threshold can depend on $p$; because $\alpha_n\to0$, the triangular array
eventually enters that threshold.

If the $L$ recursions are started from deterministic values and the first
$b_n$ observations are discarded, the initialization contribution after TD
burn-in is controlled by
$\exp(-c a_{\min}\alpha_n b_n)$.  Thus a sufficient replacement for
\Cref{eq:theta-star-deterministic-rate-main} is
\[
    \frac{\exp(-c a_{\min}\alpha_n b_n)}
         {\alpha_n\sqrt n}
    \longrightarrow0,
\]
with the statistic constructed from the retained segment.  

If instead, for the reported contrast,
\[
    \sqrt n\,R(\theta_{\mathrm{RR},n}-\theta^*)\to\delta\ne0,
\]
the Brownian-bridge
self-normalizer is unchanged because a linear deterministic drift cancels
under bridge centering, but the contrast endpoint is shifted by $\delta$.
The limiting law in \Cref{cor:self-normalized-theta-star} then becomes
noncentral, so a confidence region constructed as if the residual shift were
zero no longer has the stated coverage.
 \section{Numerical Experiments}
\label{sec:experiments}

\subsection{Setup, Methods, and Inferential Targets}

We use two finite-state policy-evaluation benchmarks to examine three
implications of the theory: coverage at the appropriate stationary target,
reduction of the fixed-stepsize target shift by RR, and the change in
remainder behavior when the stepsize decreases across horizons.  FrozenLake
provides a structured episodic environment with two choices of feature
dimension,
whereas Garnet permits controlled variation in feature dimension and mixing.
Within each replication, all methods and RR levels use the same simulated
trajectory.  Detailed environment generation, feature construction, exact
target computation, pilot tuning, and additional results are given in Online
Appendix 2.  Code and the archived result summaries used to regenerate the
reported figures and tables are available at
\url{https://github.com/MinZenggit/SN-TD}.

The FrozenLake study uses slippery $8\times8$ maps, feature dimensions
$d\in\{4,10\}$, and base stepsizes
\[
    \alpha\in\{0.005,0.01,0.02,0.04,0.06\}.
\]
The Garnet study uses 100-state MDPs, dimensions $d\in\{5,10\}$, paired fast-
and slow-mixing transition kernels, and
\[
    \alpha\in\{0.005,0.01,0.02,0.04\}.
\]
For the fixed-stepsize comparisons, FrozenLake uses $\gamma=0.99$ and a TD
burn-in of $\lceil2000/\alpha\rceil$, whereas Garnet uses $\gamma=0.95$ and a
200,000-step TD burn-in.  After a terminal transition in FrozenLake, the
process resets to the designated start state, while the terminal feature
remains in the TD update; this convention defines the continuing Markov
chain used for stationary analysis.
Each cell combines ten environments, three feature draws per environment, and
ten trajectories per feature draw, for 300 trajectories at the terminal
horizon $n=10^6$.  The environment state is initialized from its exact
invariant distribution.

All reported contrasts are scalar and fixed before simulation.  For
FrozenLake, $R$ is the normalized feature row of the designated start state;
for Garnet, it is the normalized feature row of state zero.  The corresponding
values of $R\theta_\alpha$, $R\theta_{\mathrm{RR},\alpha}$, and $R\theta^*$
are computed from the exact finite-state linear systems.

\begin{figure}[t]
\centering
\includegraphics[width=0.92\linewidth]{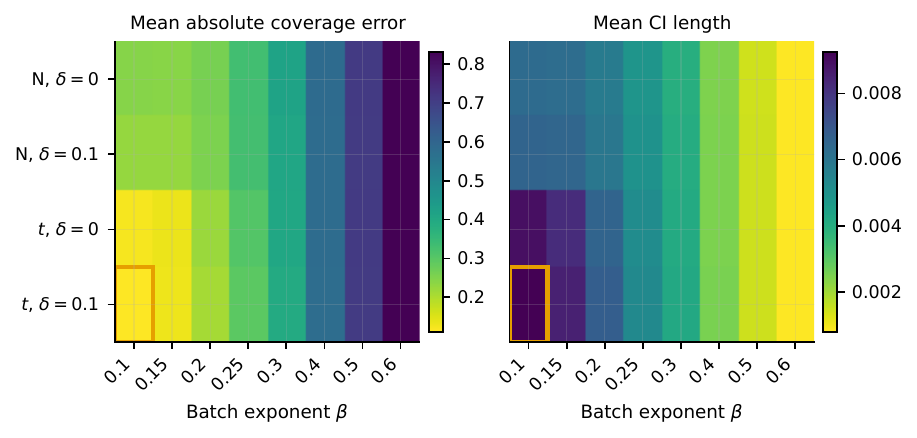}
\caption{FrozenLake pilot-only sensitivity of Huo batch-means inference.  The
left panel gives the mean absolute coverage error across the five pilot
stepsizes, and the right gives mean interval length.  Row labels use N for the
normal critical value and $t$ for the Student critical value.  Each candidate uses
1,200 pilot intervals; all pilot maps, features, and trajectories are disjoint
from evaluation.  The orange box marks the Huo-BM-pilot rule selected in the
pilot study and fixed for the evaluation runs ($\beta=0.1$, deletion fraction
$0.1$, Student critical value).}
\label{fig:huo-pilot-tuning}
\end{figure}

For every base stepsize, the RR recursions use $(\alpha,2\alpha)$ and weights
$(2,-1)$.  We compare self-normalized RR (RR--SN), two implementations of the
batch-means procedure of \citet{HuoChenXie2024AAAI}, and self-normalization
without RR (No-RR SN).  Huo-BM-0.3 uses batch exponent $0.3$, no deletion, and
a normal critical value; Huo-BM-pilot uses exponent $0.1$, deletion fraction
$0.1$, and a Student critical value.
Accordingly, stationary-target coverage refers to coverage of
$R\theta_{\mathrm{RR},\alpha}$ for RR--SN and the two Huo procedures, and of
$R\theta_\alpha$ for No-RR SN.  Coverage of $R\theta^*$ is reported
separately.  All reported intervals are two-sided 95\% confidence intervals
for scalar contrasts.  Environment-cluster intervals use
the ten environments as resampling units and are descriptive finite-design
summaries.
The exponent in Huo-BM-0.3 reproduces the scaling $K\asymp n^{0.3}$ used
in the trajectory-length experiment of \citet{HuoChenXie2024AAAI}; it is a
literature-based benchmark rather than a universally preferred rule.
Independent, benchmark-specific pilots used environments, features, and
trajectories disjoint from evaluation.  Both pilots prioritized worst-case
coverage error, with interval length used only as a tie-breaker, and selected
the same rule, which was then fixed for evaluation.  \Cref{fig:huo-pilot-tuning}
shows the FrozenLake pilot grid, and Online Appendix 2 gives the exact
benchmark-specific ranking criteria.  The accompanying reproducibility
repository records the selected rules, simulation seeds, and regeneration
scripts.

\FloatBarrier
\subsection{Fixed-Stepsize Coverage}

\begin{figure}[t]
\centering
\includegraphics[width=0.95\linewidth]{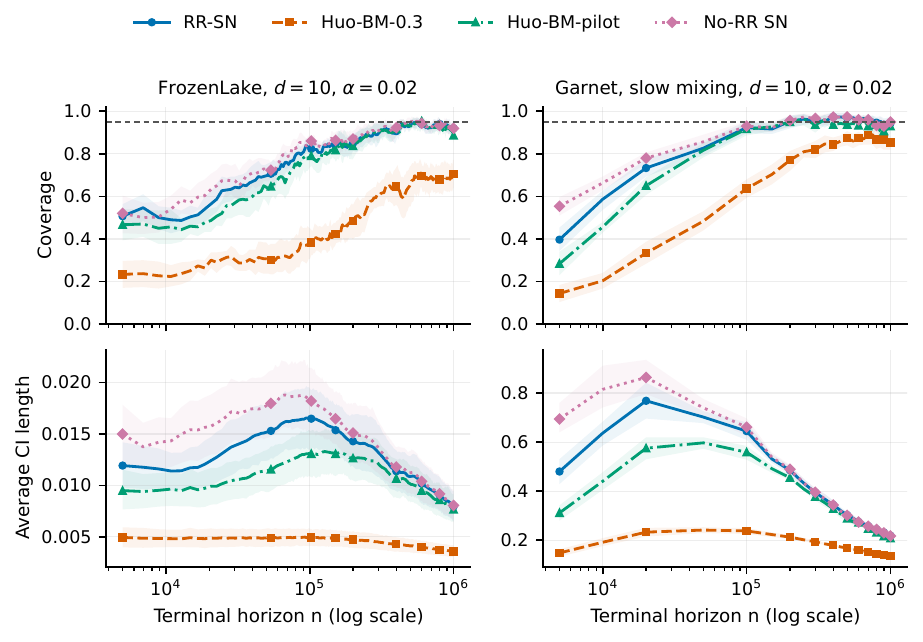}
\caption{Coverage of each method's stationary target (top) and average 95\%
confidence-interval length (bottom) as functions of the reporting horizon at
$\alpha=0.02$.  The left panels show FrozenLake with $d=10$; the right panels
show the slow-mixing Garnet design with $d=10$.  Shaded regions are 95\%
environment-cluster intervals based on ten maps or MDPs, and the horizontal
dashed line marks nominal coverage.  Coverage is pointwise in the horizon.}
\label{fig:methods-by-horizon}
\end{figure}

We report coverage along the trajectory separately from sensitivity to the
fixed stepsize.  \Cref{fig:methods-by-horizon} tracks coverage and interval
length over the reporting horizon at the common intermediate stepsize
$\alpha=0.02$.  Each curve measures coverage of the stationary target assigned
to that method; in particular, the No-RR target differs from the RR and Huo
targets.  In both benchmarks, the empirical coverage of the self-normalized
procedures stabilizes gradually as the reporting horizon increases.  The
shorter Huo-BM-0.3 intervals coincide with
substantial undercoverage, especially in the higher-dimensional FrozenLake
design.

\begin{figure}[t]
\centering
\includegraphics[width=0.92\linewidth]{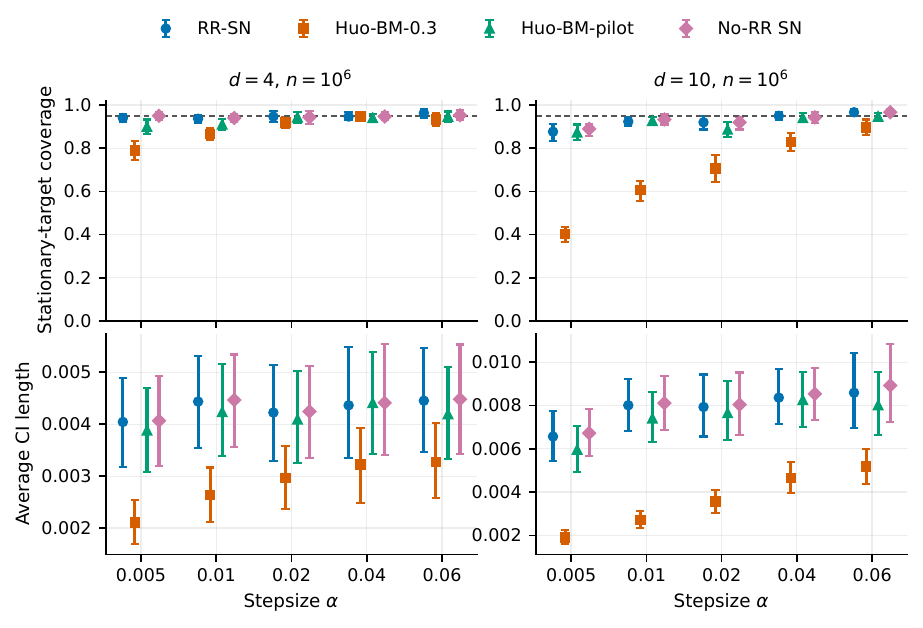}
\caption{FrozenLake coverage of each method's stationary target (top) and
average 95\% confidence-interval length (bottom) at $n=10^6$.  The five
stepsizes are displayed as discrete categories without connecting lines;
points are horizontally offset by method.  Error bars are descriptive 95\%
map-cluster intervals based on ten maps.}
\label{fig:frozenlake-main-comparison}
\end{figure}

On FrozenLake, RR--SN coverage of the RR stationary target ranges from 0.937
to 0.963 for $d=4$.  For $d=10$, the two larger stepsizes are close to
nominal, whereas the three smaller stepsizes remain under-covered at
$n=10^6$, with coverage $0.877$, $0.923$, and $0.920$.  The smaller-stepsize
$d=10$ cells also stabilize more slowly over the observed horizons.  The
present experiment does not isolate the mechanism underlying this
finite-sample pattern.  The decision to extend the FrozenLake runs to $n=10^6$
was made after examining the shorter-horizon diagnostics.  We therefore
interpret the endpoint and 201-horizon results as an extended sensitivity
analysis.
\Cref{fig:frozenlake-main-comparison} reports endpoint coverage and length over
the stepsize grid.  Huo-BM-0.3 produces shorter intervals
but substantial undercoverage, so the length difference should not be
interpreted as an efficiency gain.  Relative to Huo-BM-0.3, Huo-BM-pilot
improves FrozenLake coverage with longer intervals.  Online Appendix 2 reports
all endpoint cells and the full
201-horizon RR--SN paths.

Across the 16 Garnet cells, RR--SN coverage ranges from 0.927 to 0.963, with
every MDP-cluster interval containing 0.95.  In the paired design, slower
mixing is associated primarily with wider intervals and slower finite-horizon
coverage stabilization, as illustrated by the right panels of
\Cref{fig:methods-by-horizon}: the average RR--SN interval is approximately
five times longer under slow mixing, while the corresponding mean coverages
are about $0.95$ and $0.94$.  Huo-BM-0.3 undercovers in several cells.
Huo-BM-pilot is closer to nominal but depends on a batch rule selected on an
independent pilot.  Because the methods do not attain identical coverage,
their interval lengths provide a descriptive rather than coverage-matched
comparison.  Complete
RR--SN endpoint cells and all-method coverage paths are reported in Online Appendix 2.

The isolated implementation benchmark in
\Cref{tab:frozenlake-n1m-computation} complements the asymptotic memory count
in \Cref{sec:algorithm}.  In this implementation, RR--SN maintained
$0.180$ MB of online state and ran about 8--11\% faster than the two Huo-BM
implementations.  The timing order is specific to this implementation and
software stack.

{\setlength{\tabcolsep}{4.5pt}\begin{table}[!htbp]
\centering
\footnotesize
\setlength{\tabcolsep}{2pt}
\begin{tabular}{@{}lrrrrrr@{}}
\toprule
Method & \shortstack{TD recursions\\per transition} & \shortstack{Time\\(s)} & \shortstack{Relative\\time} & \shortstack{Incremental\\RSS (MB)} & \shortstack{Online\\state (MB)} & \shortstack{Stored batch\\boundaries} \\
\midrule
RR-SN & 2 & 196.77 & 1.00 & 0.53 & 0.180 & 0 \\
Huo-BM-0.3 & 2 & 222.21 & 1.13 & 2.67 & 0.517 & 9444 \\
Huo-BM-pilot & 2 & 213.99 & 1.09 & 1.02 & 0.232 & 1974 \\
No-RR SN & 1 & 165.36 & 0.84 & 0.55 & 0.180 & 0 \\
\bottomrule
\end{tabular}
\caption{Isolated $d=10$, $n=10^6$ computational benchmark. Medians are computed over five independent process launches, each processing five paths and 201 reporting horizons. Incremental RSS is measured above the post-setup baseline.}
\label{tab:frozenlake-n1m-computation}
\end{table}
 }

\FloatBarrier
\subsection{RR Reduction of the Target Shift}

\begin{figure}[tbp]
\centering
\includegraphics[width=0.88\linewidth]{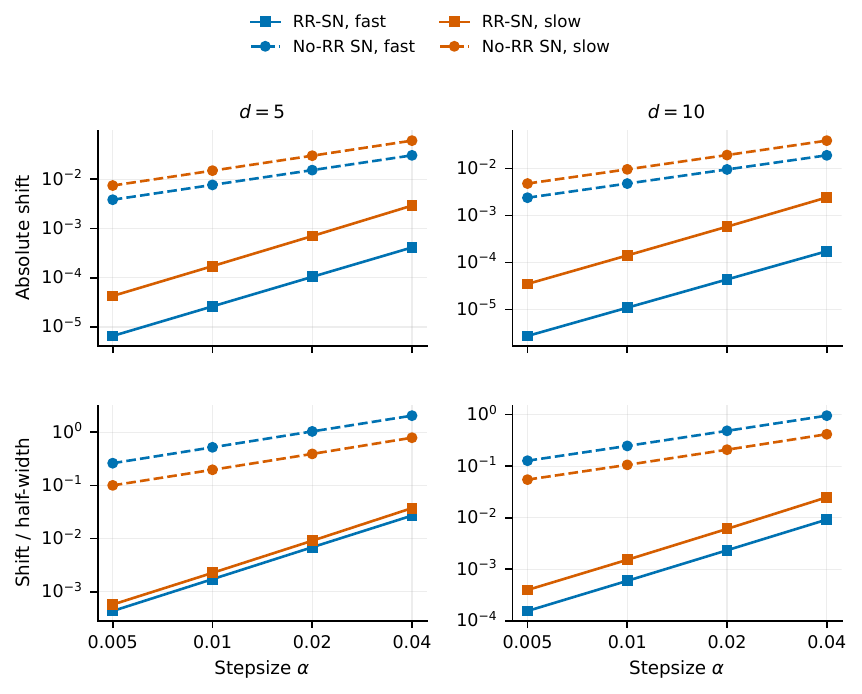}
\caption{Garnet target shift from $R\theta^*$ for RR--SN and No-RR SN.  The
top row reports the mean absolute target shift, and the bottom row divides each
replication-level target shift by its interval half-width before averaging.
The columns correspond to $d=5$ and $d=10$.
Solid lines with square markers denote RR--SN, and dashed lines with circles
denote No-RR SN.  The reduction is a finite-design comparison across the
evaluated stepsizes, dimensions, and mixing regimes.}
\label{fig:garnet-target-mismatch}
\end{figure}

To assess whether a constant-stepsize interval is also centered near the
projected Bellman solution, we report coverage of $R\theta^*$ in addition to
stationary-target coverage.  In the Garnet study, No-RR SN has mean coverage
about $0.95$ for $R\theta_\alpha$ but covers $R\theta^*$ only about $0.81$ of
the time.  The first-order RR combination reduces the mean absolute target
shift from approximately $1.75\times10^{-2}$ to $4.9\times10^{-4}$, about a
36-fold reduction in the evaluated Garnet designs, while leaving average
interval length essentially unchanged.  As shown
in \Cref{fig:garnet-target-mismatch}, this comparison isolates RR target
correction from self-normalized coverage
around each method's own stationary target.

\FloatBarrier
\subsection{Horizon-Indexed Diagnostics}
\label{sec:garnet-fixed-c}

\Cref{thm:theta-star-vanishing-fclt} gives separate rate conditions for the
root-$n$ residual RR target shift and the deterministic-start
contribution.  We examine these quantities across independently simulated
constant-stepsize runs with
\[
    \alpha_{n,\nu}=0.4n^{-\nu},\qquad
    \nu\in\{0.20,1/4,1/3,1/2,0.60\},
\]
at four horizons from $10^5$ to $10^6$.  The TD parameters start from zero,
while the environment starts in stationarity.  We report direct coverage
together with two remainder diagnostics because wide transient intervals can
produce conservative coverage.  The Monte Carlo mean-center diagnostic is an
indirect proxy for the initialization contribution rather than the pathwise
remainder in the proof.

For each diagnostic $D_n$, \Cref{tab:garnet-fixed-c-slopes} reports the fitted
slope $s$ from regressing $\log D_n$ on $\log n$ across the four horizons.  Thus a
negative slope indicates decay, a slope near zero indicates an approximately
flat diagnostic over this grid, and a positive slope indicates growth.  For first-order RR, the proof-based
benchmark exponents are $1/2-2\nu$ for the root-$n$ residual target shift and
$\nu-1/2$ for the deterministic-initialization term.

The RR target-shift diagnostic follows the predicted direction across the
grid: it grows at $\nu=0.20$, is approximately flat at $\nu=1/4$, and decays
at $\nu=1/3$.  At the interior value $\nu=1/3$, its fitted slope is $-0.168$,
close to the benchmark $-1/6$.  As a separate bias-order control reported in
Online Appendix 2, the corresponding No-RR slope is $0.166$, close to
$+1/6$.  The initialization proxy decays at $\nu=1/3$, does not decay at
$\nu=1/2$, and grows at $\nu=0.60$.  These four-horizon fits assess the
sufficient remainder scales over the evaluated grid; they do not identify
necessary rate-window boundaries.  Online Appendix 2 reports the cell-level
RR versus No-RR bias-order curves and the endpoint coverage table.

\begin{table}[htbp]
\centering
\footnotesize
\setlength{\tabcolsep}{4pt}
\begin{tabular}{@{}llrrrr@{}}
\toprule
& & \multicolumn{2}{c}{Root-$n$ RR target shift}
& \multicolumn{2}{c}{Initialization proxy} \\
\cmidrule(lr){3-4}\cmidrule(l){5-6}
$\nu$ & Position & Fitted $s$ & $1/2-2\nu$ & Fitted $s$ & $\nu-1/2$ \\
\midrule
0.20 & Below window   &  0.094 &  0.100 & -0.307 & -0.300 \\
1/4  & Lower boundary & -0.004 &  0.000 & -0.254 & -0.250 \\
1/3  & Interior       & -0.168 & -0.167 & -0.167 & -0.167 \\
1/2  & Upper boundary & -0.500 & -0.500 &  0.136 &  0.000 \\
0.60 & Above window   & -0.660 & -0.700 &  0.387 &  0.100 \\
\bottomrule
\end{tabular}
\caption{Fixed-$c$ Garnet rate diagnostics for
Theorem~\ref{thm:theta-star-vanishing-fclt}.  Each fitted entry is the slope
$s$ from a regression of $\log D_n$ on $\log n$ over four horizons, averaged
over fast/slow mixing and $d\in\{5,10\}$.  Position is relative to the
sufficient first-order RR window $1/4<\nu<1/2$; the two benchmark columns give
the corresponding proof-based exponents.  Negative and positive slopes
indicate decay and growth, while a slope near zero indicates an approximately
flat trend over this grid.  The
initialization diagnostic is a Monte Carlo mean-center proxy, not the pathwise
proof remainder.}
\label{tab:garnet-fixed-c-slopes}
\end{table}
 
\begin{figure}[t]
\centering
\includegraphics[width=0.88\linewidth]{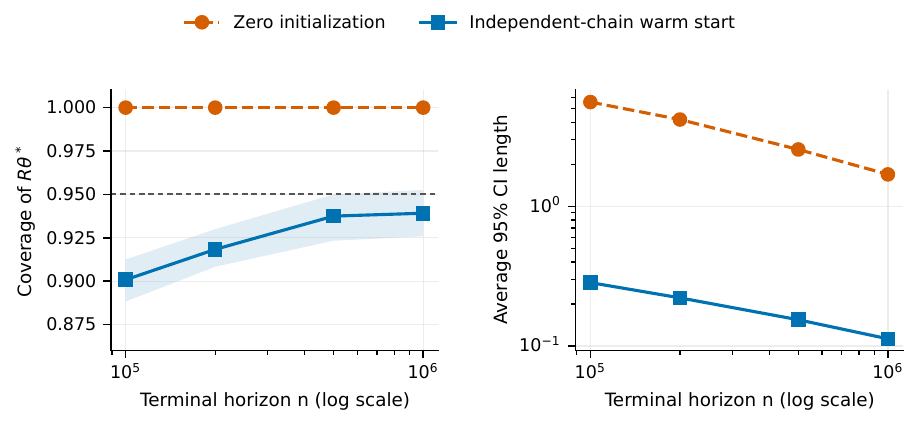}
\caption{Direct-$R\theta^*$ coverage (left) and average interval length
(right) for RR--SN at $\nu=1/3$, comparing zero initialization with the
reference independent-chain 200,000-step TD warm start.  The evaluation
trajectory begins from a separate stationary state and is shared by the two
initialization branches.  Each horizon pools 1,200 intervals over mixing
regimes and dimensions.  The shaded region is the 95\% MDP-cluster interval
for warm-start coverage; the length axis is logarithmic.}
\label{fig:garnet-fixed-c-reference-warmup-main}
\end{figure}

The independent-chain TD warm-start sensitivity in
\Cref{fig:garnet-fixed-c-reference-warmup-main} provides a direct coverage
check at the interior exponent $\nu=1/3$.  The TD parameters are first updated
for 200,000 steps on an independent stationary trajectory; the evaluation
trajectory then starts from a separate stationary state and is shared
with the zero-initialization branch.  At $n=10^6$, this warm start gives
RR--SN coverage about $0.94$, with MDP-cluster interval
$[0.926,0.953]$ and average length about $0.11$.  Zero initialization instead gives
coverage $1.000$ with average length about $1.69$.  The latter is conservative
transient widening rather than evidence of more accurate inference.  The
comparison is a finite-horizon sensitivity analysis and does not select an
optimal warm-start length.  Online Appendix 2, Table 4 reports the corresponding
horizon-by-horizon values for RR--SN and No-RR SN.

\FloatBarrier
 \section{Discussion}
\label{sec:conclusion}

Our results distinguish two inferential regimes for constant-stepsize TD.  At
a fixed stepsize, self-normalization quantifies sampling fluctuation around the
RR stationary target; interpreting the same region as inference for the
projected Bellman solution additionally requires separate control of the
residual RR target shift relative to the sampling uncertainty.  When stepsizes
are held constant within separately planned runs and decrease across
prespecified reporting horizons, the rate and initialization conditions make
the residual target shift, multiplicative remainder, and initialization effect
negligible at the root-$n$ scale, yielding direct inference for $\theta^*$.
At a fixed stepsize, however, the path law must retain both the multiplicative
TD component and the dependence among RR levels induced by the shared
trajectory.

The experiments illustrate the corresponding finite-horizon behavior.
RR--SN attains near-nominal stationary-target coverage across the evaluated
Garnet designs, whereas the smaller-stepsize, higher-dimensional FrozenLake
cells remain under-covered and stabilize more slowly over the observed
horizons.  This pattern is consistent with poorer conditioning in the
higher-dimensional design, although the experiments do not isolate its effect
from other design differences.  In the paired Garnet design, slower mixing is
associated primarily with wider intervals, while RR substantially reduces the
measured target shift without materially changing interval length.  The
horizon-indexed diagnostics are consistent with the predicted directions of
the sufficient remainder scales over the evaluated horizon grid, including
the RR/No-RR slope reversal at the interior rate $\nu=1/3$.  At that rate, the
warm-start comparison shows that initialization can materially affect
finite-horizon direct-$\theta^*$ coverage and interval length.  The runtime and
memory results are implementation-specific rather than complexity lower
bounds.

The admissible stepsize thresholds and exponent window are sufficient proof
conditions rather than tuning prescriptions.  In practice, stability,
conditioning, mixing, and sensitivity across reporting horizons provide useful
diagnostics, while adaptive selection of the stepsize, RR design, or rate
exponent with valid post-selection inference remains an open problem.

The theory is fixed-dimensional and linear, and assumes bounded, stationary,
uniformly geometrically ergodic Markov sampling together with a prespecified
contrast having nondegenerate limiting covariance.  The experiments consider
scalar contrasts, although the theory allows fixed multivariate contrasts.
Extending the framework to nonlinear function approximation, adaptive
policies, learner-dependent sampling, or nonstationary data starts would
require additional stability and dependence arguments.  Likewise, the present
confidence statements are pointwise at a prespecified horizon; optional
stopping, simultaneous inference across horizons, and confidence sequences
require a different inferential framework.  Within this scope, one-pass
inference is available at a prespecified horizon, centered on
$\theta_{\mathrm{RR},\alpha}$ at a fixed stepsize and on $\theta^*$ under the
horizon-indexed rate and initialization conditions.
 
\acks{Xiaofeng Shao would like to acknowledge partial support from the startup
fund provided by the Department of Statistics and Data Science at Washington
University in St. Louis. The authors declare that they have no competing
interests.}

\appendix
\section{Admissible Stepsizes and Assumption Map}
\label{app:threshold-guide}

This appendix collects the sufficient stepsize constants used in the proofs;
practical tuning is beyond their intended role.  An RR node $a_\ell$ is a
fixed positive multiplier.  The base stepsize is $\alpha$, and the $\ell$th
recursion uses the level-specific stepsize $\alpha_\ell=a_\ell\alpha$.
Per-recursion thresholds apply to the $\alpha_\ell$ values, whereas base-step
thresholds apply directly to $\alpha$.

Under \Cref{ass:hurwitz}, let $Q$ be the unique positive-definite solution of
\begin{equation}
    \barA^\top Q+Q\barA=\Id,
    \label{eq:main-lyapunov-Q}
\end{equation}
and write $q_{\min}=\lambda_{\min}(Q)$,
$q_{\max}=\lambda_{\max}(Q)$, $\kappa_Q=q_{\max}/q_{\min}$, and
$\eta_Q=q_{\max}^{-1}$.  Set
\[
    K_Q:=\sup_y\|Q^{1/2}\calA(y)Q^{-1/2}\|,
    \quad
    M_Q:=\frac{2C_0K_Q\rho}{1-\rho},
    \quad
    \ell_Q:=\max\left\{1,\left\lceil\frac{2M_Q}{\eta_Q}\right\rceil\right\}.
\]
For $\ell=\ell_Q$, define
\[
    \delta_\ell(a):=(1+aK_Q)^\ell-1,
    \quad
    r_\ell(a):=(1+aK_Q)^\ell-1-a\ell K_Q,
    \quad
    \psi_p(s):=(1+s)^p-1-ps.
\]
The per-recursion stability threshold in \Cref{ass:stepsize} is
\begin{align}
    \alpha_{\mathrm{root},p}
    &:={}
    \sup\left\{
       a>0:
       p r_{\ell_Q}(s)+\psi_p\{\delta_{\ell_Q}(s)\}
       \le \frac{ps\eta_Q\ell_Q}{8}
       \ \text{for every }0<s\le a
    \right\}, \notag\\
    \alpha_{\mathrm{stab},p}
    &:={}
    \min\left\{
       \alpha_{\mathrm{root},p},
       \frac{4}{p\eta_Q\ell_Q}
    \right\}.
    \label{eq:main-alpha-stab}
\end{align}
It is positive because the left-hand side is $O(s^2)$ as $s$ decreases to
zero.  Online Appendix 1 proves the resulting block-contraction
bound and gives a more conservative closed-form alternative.

For a fixed RR design, let
$a_{\max}:=\max_{1\leq\ell\leq L}a_\ell$.  The base threshold used for all
fixed-stepsize results is
\begin{equation}
    \alpha_{\mathrm{fix},p}
    :=\frac{\alpha_{\mathrm{stab},p}}{a_{\max}}.
    \label{eq:main-alpha-fix}
\end{equation}
Thus $\alpha\leq\alpha_{\mathrm{fix},p}$ ensures that every RR level uses a
level-specific stepsize satisfying
$a_\ell\alpha\leq\alpha_{\mathrm{stab},p}$.

The horizon-indexed argument uses two further per-recursion thresholds.
The threshold $\alpha_{\mathrm{mom},p}>0$ guarantees
\begin{equation}
    \sup_{0<\beta\leq
      \min\{\alpha_{\mathrm{stab},p},\alpha_{\mathrm{mom},p}\}}
    \beta^{-1/2}
    \|\theta_0^{\beta,\circ}-\theta^*\|_{L^p}<\infty,
    \label{eq:main-alpha-mom-property}
\end{equation}
and $\alpha_{\mathrm{bias},m}>0$ guarantees an order-$m$ expansion
\begin{equation}
    \left\|
      \theta_\beta-\theta^*-
      \sum_{j=1}^{m}\beta^j\Delta_j
    \right\|
    \leq C_m\beta^{m+1},
    \qquad
    0<\beta\leq
      \min\{\alpha_{\mathrm{stab},p},\alpha_{\mathrm{bias},m}\}.
    \label{eq:main-alpha-bias-property}
\end{equation}
Online Appendix 1 constructs these constants where the corresponding external
moment and bias-expansion results are first used.  In particular,
$\alpha_{\mathrm{mom},p}$ depends on the drift and minorization quantities in
that moment theorem; the pair $(C_0,\rho)$ alone does not numerically determine
all of those quantities.

Define the remaining base thresholds by
\begin{equation}
    \alpha_{\mathrm{RR},m}
    :=\frac{\alpha_{\mathrm{bias},m}}{a_{\max}},
    \qquad
    \alpha_{\mathrm{adm},p,m}
    :=\min\left\{
       \alpha_{\mathrm{fix},p},
       \frac{\alpha_{\mathrm{mom},p}}{a_{\max}},
       \alpha_{\mathrm{RR},m}
    \right\}.
    \label{eq:main-alpha-adm}
\end{equation}
The subscript $m$ records the order of the finite bias expansion.  Although the
corresponding stepsize restriction used in Online Appendix 1 does not
numerically depend on $m$, the expansion coefficients and remainder constant
generally do.

\begin{table}[t]
\centering
\small
\setlength{\tabcolsep}{3.2pt}
\begin{tabular}{@{}lp{2.1cm}p{6.2cm}l@{}}
\toprule
Threshold & Scale & Role & Main use \\
\midrule
$\alpha_{\mathrm{stab},p}$
& per recursion & $L^p$ block contraction and stationarity
& fixed-stepsize FCLTs \\
$\alpha_{\mathrm{fix},p}$
& base step & stability at every RR level
& \Cref{ass:stepsize} \\
$\alpha_{\mathrm{mom},p}$
& per recursion & $O(\sqrt\alpha)$ stationary moment
& \Cref{sec:theta-star} \\
$\alpha_{\mathrm{bias},m}$
& per recursion & order-$m$ stationary-bias expansion
& RR bias bound \\
$\alpha_{\mathrm{RR},m}$
& base step & bias-expansion admissibility at every RR level
& \Cref{sec:theta-star} \\
$\alpha_{\mathrm{adm},p,m}$
& base step & combined stability, moment, and bias conditions at all RR levels
& \Cref{ass:horizon-indexed} \\
\bottomrule
\end{tabular}
\caption{Stepsize thresholds used in the analysis.}
\label{tab:main-threshold-map}
\end{table}

\FloatBarrier
\section{From the Functional Limit to Self-Normalized Inference}
\label{app:self-normalization-proof}

This appendix gives the common argument that converts the two FCLTs in the
paper into pivotal confidence regions.  It also makes explicit why the method
does not require a separate long-run covariance estimator.

\begin{lemma}[Brownian-bridge matrix]
\label[lemma]{lem:main-brownian-bridge}
Let $W_q$ be a standard $q$-dimensional Brownian motion and
$\bar W_q(r)=W_q(r)-rW_q(1)$.  Then
\[
    K_q:=\int_0^1\bar W_q(r)\bar W_q(r)^\top\,dr
\]
is positive definite almost surely, and
$W_q(1)^\top K_q^{-1}W_q(1)$ has a continuous distribution.
\end{lemma}

\begin{proof}
If $K_q$ is singular, continuity implies that
$v^\top\bar W_q(r)=0$ for all $r\in[0,1]$ and some nonzero $v$.  At any fixed
distinct times $0<r_1<\cdots<r_q<1$, the matrix
$[\bar W_q(r_1),\ldots,\bar W_q(r_q)]$ is therefore singular.  Its
vectorization is a nondegenerate Gaussian vector because the covariance matrix
of a scalar Brownian bridge at distinct interior times is positive definite.
The determinant-zero set has Lebesgue measure zero, proving
$K_q\succ0$ almost surely.  Moreover, $W_q(1)$ is independent of the bridge.
Conditional on $K_q$, the quadratic form is a positive weighted sum of
independent chi-squared variables and has a continuous distribution.  Averaging
over $K_q$ proves continuity of the unconditional law.
\end{proof}

\begin{proposition}[Self-normalization transfer]
\label[proposition]{prop:main-sn-transfer}
Suppose a sequence of processes, possibly arising from a triangular array,
satisfies
\[
    C_n(r):=n^{-1/2}\sum_{t=1}^{\lfloor nr\rfloor}(X_{t,n}-\theta_n)
    \dto \Omega^{1/2}W_d(r)
\]
in $D([0,1],\R^d)$, where $\Omega$ is positive semidefinite.  Let
$R\in\R^{q\times d}$ have full row rank and
$\Gamma:=R\Omega R^\top\succ0$.  With
$S_{s,n}=\sum_{t=1}^sX_{t,n}$, $\bar X_n=S_{n,n}/n$, and
\[
    \widehat V_n
    :=\frac1{n^2}\sum_{s=1}^n
      (S_{s,n}-s\bar X_n)(S_{s,n}-s\bar X_n)^\top,
\]
we have $\Pr(R\widehat V_nR^\top\succ0)\to1$.  Under
$H_0:R\theta_n=c$, define
\[
    T_n:=
    \begin{cases}
    n(R\bar X_n-c)^\top(R\widehat V_nR^\top)^{-1}(R\bar X_n-c),
      & R\widehat V_nR^\top\succ0,\\
    0, & \text{otherwise}.
    \end{cases}
\]
Then $T_n\dto W_q(1)^\top K_q^{-1}W_q(1)$.
\end{proposition}

\begin{proof}
Uniformly in $r$,
\[
    n^{-1/2}\sum_{t=1}^{\lfloor nr\rfloor}(X_{t,n}-\bar X_n)
    =C_n(r)-rC_n(1)+o_p(1).
\]
Because the limiting paths are continuous, the continuous mapping theorem,
including the Riemann-sum map, gives
\[
    \widehat V_n
    \dto
    \Omega^{1/2}
    \left\{\int_0^1\bar W_d(r)\bar W_d(r)^\top\,dr\right\}
    \Omega^{1/2}.
\]
Define
\[
    \widetilde W_q(r):=\Gamma^{-1/2}R\Omega^{1/2}W_d(r).
\]
Its covariance is $\min(r,s)\Id_q$, so it is a standard $q$-dimensional
Brownian motion.  Joint convergence of the endpoint and bridge yields
\[
    R\widehat V_nR^\top
    \dto
    \Gamma^{1/2}
    \left\{\int_0^1
      \overline{\widetilde W}_q(r)
      \overline{\widetilde W}_q(r)^\top\,dr
    \right\}
    \Gamma^{1/2}.
\]
The limiting matrix is positive definite almost surely by
\Cref{lem:main-brownian-bridge}; hence the positive-definite event has
probability tending to one.  Under the null,
$\sqrt n(R\bar X_n-c)\dto\Gamma^{1/2}\widetilde W_q(1)$ jointly with the
normalizer.  Matrix inversion is continuous on the positive-definite cone, and
the two factors $\Gamma^{1/2}$ cancel from the quadratic form.  Renaming
$\widetilde W_q$ as $W_q$ proves the result.
\end{proof}

\begin{proof}[Proof of \Cref{cor:random-scaling}]
Apply \Cref{prop:main-sn-transfer} with
$X_{t,n}=\vartheta_t^\alpha$,
$\theta_n=\theta_{\mathrm{RR},\alpha}$, and
$\Omega=\Omega_{\vartheta,\alpha}$.  The continuous limiting distribution in
\Cref{lem:main-brownian-bridge} justifies use of its
$(1-\eta)$ quantile and proves the stated coverage.
\end{proof}

\begin{proof}[Proof of \Cref{cor:theta-star-rs}]
Apply \Cref{prop:main-sn-transfer} to the deterministic-initialization FCLT in
\Cref{thm:theta-star-vanishing-fclt}, with
$X_{t,n}=\vartheta_{t,n}$, $\theta_n=\theta^*$, and $\Omega=\Omega_0$.
The same Brownian functional therefore applies in the horizon-indexed regime.
\end{proof}

\bibliographystyle{plainnat}

\end{document}